\documentclass[sigconf]{acmart}

\AtBeginDocument{%
  }

\copyrightyear{2022}
\acmYear{2022}
\setcopyright{acmcopyright}
\acmConference[AIES'22]{Proceedings of the 2022 AAAI/ACM Conference on AI, Ethics, and Society}{August 1--3, 2022}{Oxford, United Kingdom}
\acmBooktitle{Proceedings of the 2022 AAAI/ACM Conference on AI, Ethics, and Society (AIES'22), August 1--3, 2022, Oxford, United Kingdom}
\acmPrice{15.00}
\acmDOI{10.1145/3514094.3534198}
\acmISBN{978-1-4503-9247-1/22/08}

\usepackage{mathtools} %
\usepackage{booktabs} %
\usepackage{tikz} %

\usepackage{enumitem}
\usepackage{amsmath,latexsym,amsthm}
\usepackage{algorithmic,algorithm}
\usepackage{caption}
\usepackage{subcaption}
\usepackage{balance}
\newcommand{\hideh}[1]{}
\newcommand\nonumberthis{\nonumber\refstepcounter{equation}}

\newtheorem{assumption}{Assumption}

\newtheorem*{proposition*}{Proposition}
\newtheorem{lemma}{Lemma}
\newtheorem{theorem}{Theorem}
\newtheorem{theorem*}{Theorem}

\newtheorem{definition}{Definition}

\DeclarePairedDelimiterX{\infdivx}[2]{(}{)}{%
  #1\;\delimsize\|\;#2%
}
\newcommand{\infdiv}{\cD\infdivx}
\newcommand{\infdivcvar}{\cD_{\text{CVaR}}\infdivx}
\newcommand{\divcvar}{\cD_{\text{CVaR}}}

\global\long\def\reals{\mathbf{R}}

\global\long\def\E{\mathbb{E}}

\global\long\def\U{\mathcal{U}}

\global\long\def\R{\mathcal{R}}
\global\long\def\V{V}
\global\long\def\X{X}

\global\long\def\ouralg{\texttt{ROPE}}
\global\long\def\divalg{\texttt{JointDRO}}

\global\long\def\stdalg{\texttt{Standard}}

\DeclareMathOperator*{\argmin}{arg\,min}

\def\to{{\,\rightarrow\,}}

\mathchardef\mhyphen="2D

\newcommand{\vertiii}[1]{{\left\vert\kern-0.25ex\left\vert\kern-0.25ex\left\vert #1
    \right\vert\kern-0.25ex\right\vert\kern-0.25ex\right\vert}}

\def\cA{\mathcal{A}}

\def\cD{\mathcal{D}}

\def\cN{\mathcal{N}}

\def\cP{\mathcal{P}}

\def\cR{\mathcal{R}}
\def\cS{\mathcal{S}}

\def\cU{\mathcal{U}}

\begin{document}

\title{Towards Robust Off-Policy Evaluation via Human Inputs}

\author{Harvineet Singh}
\authornote{Work done as a summer fellow at the Center for Research on Computation and Society, Harvard University. Correspondence at hs3673@nyu.edu}
\affiliation{%
  \institution{New York University}
  \city{New York City}
  \state{New York}
  \country{USA}
}

\author{Shalmali Joshi}
\author{Finale Doshi-Velez}
\author{Himabindu Lakkaraju}
\affiliation{%
  \institution{Harvard University}
  \city{Cambridge}
  \state{Massachusetts}
  \country{USA}
}

\renewcommand{\shortauthors}{Singh et al.}

\begin{abstract}
  Off-policy Evaluation (OPE) methods are crucial tools for evaluating policies in high-stakes domains such as healthcare, where direct deployment is often infeasible, unethical, or expensive. When deployment environments are expected to undergo changes (that is, dataset shifts), it is important for OPE methods to perform robust evaluation of the policies amidst such changes. Existing approaches consider robustness against a large class of shifts that can arbitrarily change any observable property of the environment. This often results in highly pessimistic estimates of the utilities, thereby invalidating policies that might have been useful in deployment. In this work, we address the aforementioned problem by investigating how domain knowledge can help provide more realistic estimates of the utilities of policies. We leverage human inputs on which aspects of the environments may plausibly change, and adapt the OPE methods to only consider shifts on these aspects. Specifically, we propose a novel framework, Robust OPE (ROPE), which considers shifts on a subset of covariates in the data based on user inputs, and estimates worst-case utility under these shifts. We then develop computationally efficient algorithms for OPE that are robust to the aforementioned shifts for contextual bandits and Markov decision processes. We also theoretically analyze the sample complexity of these algorithms. Extensive experimentation with synthetic and real world datasets from the healthcare domain demonstrates that our approach not only captures realistic dataset shifts accurately, but also results in less pessimistic policy evaluations.
\end{abstract}



\maketitle

\section{Introduction}
Off-policy evaluation~(OPE) refers to the task of estimating the expected utility of a decision-making policy without having to deploy the policy~\cite{levine2020offline}. Such an ability is critical for vetting policies in high-stakes decision problems such as in healthcare \citep{gottesman2018evaluating}, where deploying a policy directly is often risky or unethical. Therefore, we must rely on existing data collected from alternate policies deployed possibly in a different environment. Accurate evaluation of a policy is important so that stakeholders can identify beneficial policies and discard the harmful ones.

In real world applications, OPE is a challenging task since the deployment environments often undergo changes (i.e., dataset shifts). It is critical for OPE methods to evaluate policies in a way that is robust to these changes. Prior work has proposed different solutions to address this problem. While some approaches address the scenario where potential shifts in the data are fully known in advance~\cite{kato2020off,killian2020counterfactually,zhang2020invariant}, others focus on the case where there is little to no knowledge of potential shifts in advance~\cite{subbaswamy2021evaluating,li2021evaluating} since this is more common in real world applications. Prior works that focus on robust OPE under unseen dataset shifts predominantly model the shifts by considering bounded perturbations to the joint distribution of the data~\cite{si2020distributional,hatt2021generalizing,zhou2021tabular}, inspired by adversarial machine learning literature. However, prior research has also demonstrated that such shifts can be overly conservative, and often result in pessimistic estimates of policy utilities~\cite{petrik2019beyond, duchi2019distributionally}. Moreover, accounting for many irrelevant shifts may result in poor performance on shifts of interest. For instance, consider adversarial training methods that perturb training data in an $\ell_p$-norm ball and minimize the worst-case error across the perturbations \citep{goodfellow2014explaining,madry2018towards,sinha2018certifying}. Existing work shows that robustness to shifts modeled as $\ell_p$ perturbations does not transfer well to real world shifts \citep{taori2020measuring} or even to other $\ell_p$ norms \citep{maini20union}. Such methods can degrade performance on train distribution \citep{raghunathan2020understanding} and lead to degenerate solutions \citep{hu2018does}.

\begin{figure*}[htbp!]
\centering
\begin{minipage}{0.33\textwidth}
    \centering
    \includegraphics[scale=0.25]{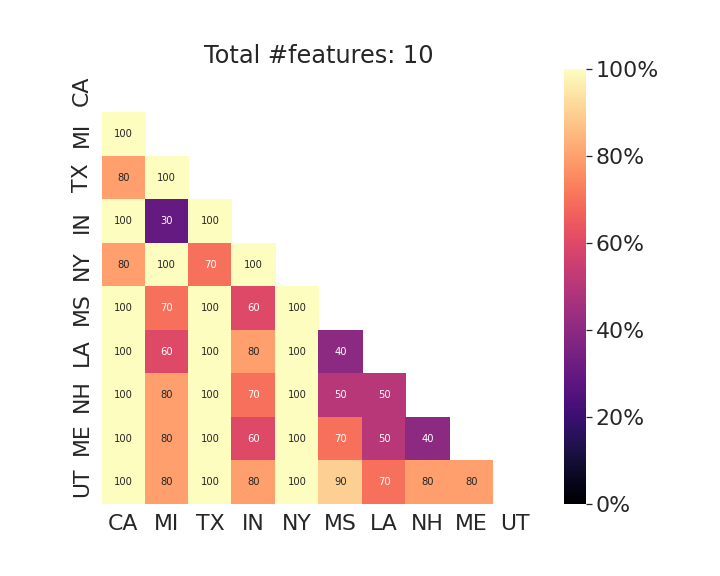}
    US Census data (Adult Income)
\end{minipage}%
\begin{minipage}{0.33\textwidth}
    \centering
    \includegraphics[scale=0.25]{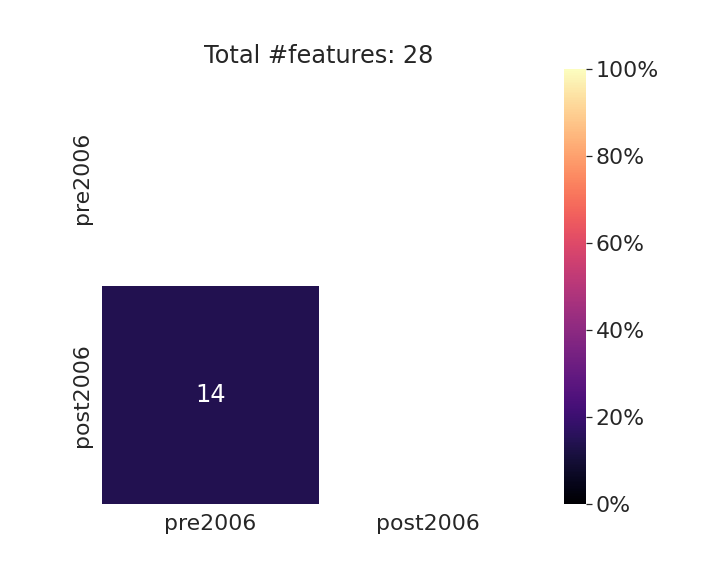}
    Loan data (SBA)
\end{minipage}%
\begin{minipage}{0.33\textwidth}
    \centering
    \includegraphics[scale=0.25]{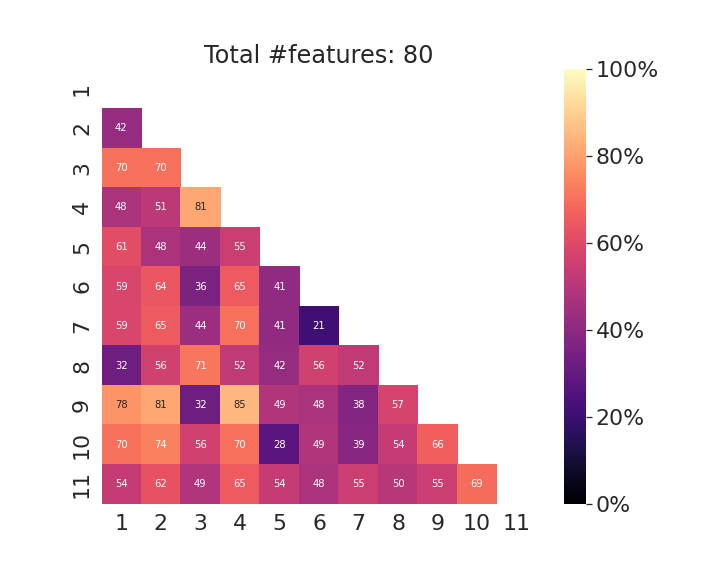}
    Medical data (eICU)
\end{minipage}
\caption{\textbf{Prevalence of shifts on subset of features.} Plots show the percentage of features that shift across domains within three different datasets (US Census, Loan, Medical). More details are given in Appendix \ref{app:shift_detect_app}. The three datasets have 10, 2, and 11 domains, respectively. The x and y-axes show the different domains (10 US state codes in US Census, 2 time periods in Loan, and 11 hospitals in Medical data). Each cell denotes the percentage of the total features (shown in each subtitle) that shift for the corresponding pair of domains. We detect and count feature shifts using conditional independence tests \citep{kulinski2020feature}, where we check if $P_\text{train}(X_i | X\setminus X_i)=P_\text{test}(X_i | X\setminus X_i)$. Firstly, plots show that distribution shifts are quite prevalent. Secondly, we observe that only a few domain pairs in US Census undergo shift in all features (value of 100\% in the leftmost plot). In all other pairs, only a fraction of the features shift (similar to Loan and Medical data). This shows that shifts on feature subsets are prevalent in the real world datasets and methods that assume shifts in \textit{all} features are aiming for robustness to unrealistic shifts.}
\label{fig:count_shifts}
\end{figure*}

While prior research has often considered bounded perturbations to the joint distribution of the data, this is quite uncommon in practice and does not represent realistic dataset shifts~\cite{subbaswamy2021evaluating,taori2020measuring}. To illustrate, we plotted the percentage of features that shift in three different real world datasets comprising of census, loan, and medical data (See Figure \ref{fig:count_shifts}). For each of these datasets, we often observe that only a subset of the covariates (and not the joint distribution of the data) undergo a change in case of dataset shifts. For example, only 4 (14\%) out of the 28 total features changed between pre 2006 and post 2006 loan datasets (Figure \ref{fig:count_shifts} -- middle). In order to model realistic dataset shifts, it therefore becomes important to exploit domain knowledge and inputs from human experts which can guide us to the plausible subset of features that are likely to shift. While such inputs have been incorporated in learning or evaluating classifiers that are robust to realistic dataset shifts~\cite{subbaswamy2021evaluating,li2021evaluating}, there is little to no work in the OPE literature that leverages domain knowledge and/or human inputs when modeling shifts in the data.

 In this work, we address the aforementioned challenges by investigating how domain knowledge can help with providing more realistic estimates of the utilities of policies. To this end, we leverage human inputs on which aspects of the environments may plausibly change and adapt the OPE methods to only consider shifts on these aspects. The hope is that this enables a domain expert to constrain the shifts to only the most relevant or plausible ones. Then, we leverage the framework of \textit{distributionally robust optimization} \citep[DRO,][]{duchi2018learning} for carrying out robust policy evaluation for contextual bandits and Markov decision processes. More specifically, we make the following key contributions:
 \begin{itemize}
     \item We propose a novel framework, Robust OPE (ROPE), which considers shifts on a subset of covariates in the data based on user inputs and estimates worst-case utility under these shifts.
     \item We develop computationally efficient algorithms for robust OPE via human inputs in case of contextual bandits and Markov decision processes.
     \item We theoretically analyze the sample complexity of the proposed algorithms.
     \item We carry out extensive experiments with synthetic and real world datasets from the healthcare domain. Our results demonstrate that $\ouralg$ can successfully tackle the over-conservatism of existing robust policy evaluation methods.
 \end{itemize}
Our work paves the way for modeling realistic dataset shifts in the context of off-policy evaluation in reinforcement learning. 
\section{Related work}

As robustness has been extensively studied in a variety of learning settings, we review only closely related works on approaches for handling unseen shifts, i.e., adversarial robustness, DRO and causal robustness.

\textbf{Adversarial Robustness}. Adversarial shifts modeled as $\ell_p$-norm perturbations have been widely considered to learn models robust to adversarial attacks \citep{goodfellow2014explaining,madry2018towards}. However, such methods provide limited gains in robustness to real world shifts \citep{taori2020measuring}. Recent work compensates for the non-realistic shifts by combining multiple $\ell_p$ norm balls \citep{maini20union}, considering shifts in a perceptual distance \citep{laidlaw2021perceptual}, or augmenting with additional datasets from adjacent domains \citep{taori2020measuring,miller20qamodels}. 
Similar methods have been extended to RL under perturbations to transition dynamics \citep{sinha2018certifying, pinto2017adversarial}, or horizon length and initial state distribution \citep{qi2020robust}.

\textbf{Distributionally Robust Optimization}. 
DRO generalizes adversarial shifts to perturbations in the distributions rather than data points \citep{duchi2018learning}. The primary mechanism of DRO is to specify \textit{uncertainty sets} that encode the uncertainty about potential test distributions. These sets can be defined over joint distribution of the data~\citep{ben2013robust, bertsimas2018data, blanchet2019quantifying,duchi2018learning}, marginals ~\citep{duchi2019distributionally, li2021evaluating}, or conditionals \citep{subbaswamy2021evaluating}, and are well explored for supervised learning. 
Interestingly, adversarial training can be understood as solving DRO with a Wasserstein metric-based set \citep{staib2017distributionally, sinha2018certifying}.
Applications of DRO have been explored in contextual bandits for policy learning~\citep{si2020distributional,mo2020learning,faury2020distributionally} and evaluation~\citep{kato2020off,jeong2020robust}. In robust MDPs, sets based on KL-divergence, $L_1,L_2$ and $L_\infty$ norms have been studied~\citep{nilim2005robust, iyengar2005robust}. Some approaches iteratively refine the sets with newly observed data, but the sets are still constructed using $L_1$ norm balls \citep{petrik2019beyond}. Thus, DRO methods (including adversarial training) lack ways to add domain knowledge and constrain the uncertainty sets, excluding some recent work \citep{subbaswamy2021evaluating}.

\textbf{Causal Robustness}.
Causal methods provide robustness to arbitrarily strong shifts by leveraging properties of the data generating process. 
For instance, using only features that cause the outcome leads to a robust model against arbitrary shifts in the features (under some structural assumptions), as shown in recent works \citep{rojas2018invariant, subbaswamy2019preventing, magliacane2018domain, rothenhausler2018anchor, peters2016causal}. A relaxation to bounded shifts has been proposed \citep{rothenhausler2018anchor,oberst2021regularizing,christiansen2020causal} for supervised learning, but in the special cases of additive shifts or linear Gaussian causal models. 
In contrast, we do not make parametric assumptions on the shifts. Under bounded shifts in conditional distributions, which is a broader class than what we consider, \citet{subbaswamy2021evaluating} propose methods for evaluating the performance of a given classification model. %

Importantly, there are gaps in the literature in applying these ideas beyond supervised learning. 
In RL, \citet{si2020distributional,hatt2021generalizing} perform OPE restricted to contextual bandits and \citet{zhou2021tabular} generalizes the DRO approach to MDPs. However, all of these works only consider shifts in the joint distribution and do not investigate the use of domain knowledge to restrict the uncertainty sets. 
As RL starts to be deployed in critical applications such as for mechanical ventilation in ICUs \citep{peine2021development}, faithful evaluation of RL policies under plausible data shifts is an important need. Thus, extending $\ouralg$ to RL and demonstrating its utility in realistic settings is a useful contribution.

\section{Preliminaries}
\label{sec:prelim}
We first introduce the robust evaluation framework based on distributionally robust optimization, then give necessary background on decision-making problems modelled as %
Contextual Bandits (CB) and Markov Decision Processes (MDPs). 

\textbf{Notation} We denote the random variable for all observable properties of a domain by $\V$. The outcome variable is denoted by $Y$. This can represent labels in supervised learning, or rewards and states in RL. Features are denoted by $\X$. For a subset of features $Z\subseteq X$, the remaining features are denoted by $X\setminus Z$. Train and test distributions over $\V$ are denoted by $P$ and $Q$ respectively (using the same notation for their densities). An uncertainty set w.r.t. a distribution $P$ is denoted by $\U_P$.

\subsection{Robust Evaluation using DRO} Say, we want to evaluate a decision-making model, parameterized by $\theta$, on a test distribution $Q$. For a given reward function $r(\theta, \V)$, this means, we want to find the value of the expected reward $\E_{\V\sim Q}[r(\theta, \V)]$. However, we do not know the test distribution $Q$ a priori. Robust evaluation methods (e.g. \citep{subbaswamy2021evaluating,li2021evaluating}) address this challenge by performing a worst-case evaluation of the model. Instead of finding the expected reward under $Q$, the \textit{robust} value of a model is defined as its worst-case reward across a set of distributions $\U_P$.
\begin{equation}
     \R(\theta, \U_P) {:=} \inf_{\tilde{P}\in\ \U_P}\ \E_{\V\sim \tilde{P}}[r(\theta, \V)]
     \label{eq:robust}
\end{equation}
where $\U_P$ is referred to as the uncertainty set. The key property of the solution to (\ref{eq:robust}) is that if the $\U_P$ is suitably chosen such that it contains the test distribution $Q$, then $\R(\theta, \U_P)\leq \E_{\V\sim Q}[r(\theta, \V)]$. Thus, we get the guarantee that the reward of the model on $Q$ is at least as good as what our robust evaluation finds. Naturally, a major part of this framework is the choice of uncertainty set $\U_P$. If we precisely choose $\U_P=\{Q\}$, then we recover the test set reward. But as we keep on increasing the uncertainty set, in the hope of including $Q$, we gradually get worse values due to the worst-case nature of the optimization (\ref{eq:robust}).
Next we describe a broad class of uncertainty sets popular in past work.

\textbf{Divergence-based Uncertainty Sets}. The uncertainty set $\U_P$ is all distributions lying in a $\delta$-ball around the train distribution, defined using a divergence metric $\cD$:
\begin{equation}
\label{eq:distance_set}
 \text{[Joint DRO set]} \qquad  \mathcal{U}^\text{div}_{P} := \{Q\ll P\quad \text{s.t.}\quad \infdiv{Q}{P}\leq \delta\}
\end{equation}
where $Q{\ll} P$ denotes absolute continuity i.e. $P(\V)=0$ implies $Q(\V)=0$.
An example is the set with: $$\infdivcvar{Q}{P} {=} \log{\sup_{\V\in \texttt{dom}(\V)}} \frac{Q(\V)}{P(\V)},$$ where CVaR stands for Conditional Value at Risk \citep{rockafellar2000optimization}. 

\textbf{Types of shifts} The above definition of $\divcvar$ assumes that $P$ and $Q$ may differ in distribution over all variables in $\V$, namely shifts in the \textit{joint distribution}. There are many ways to limit the distributions considered by $\mathcal{U}^\text{div}_{P}$. \textit{Covariate shift} (e.g. \citet{duchi2019distributionally}) assumes that  only the covariate distribution $P(X)$ may change at test time to $Q(X)$, while the conditional distribution $P(Y\vert X)$ remains the same. A slight generalization is \textit{subcovariate shift}, considered by us, that posits that $P(Z)$ may change for a subset of covariates $Z$ keeping the conditional distribution constant. This reduces the size of the set given by $\divcvar$ since many of the $Q$ are removed from the divergence set (\ref{eq:distance_set}) due to the constraint of matching the corresponding distributions in $P$.
We leverage human input to specify (sub)covariate shifts and propose OPE methods in reinforcement learning. Next, we provide preliminaries for OPE in two widely-studied settings in RL -- contextual bandits and MDPs.

\subsection{Contextual Bandits}
\label{sec:prelim_cb}
Here we have access to $n$ tuples $\{(Z_i, T_i, Y_i)\}_i$ collected with a known stochastic policy that applies treatment $T_i$ in context $Z_i$ and observes the corresponding outcome $Y_i$. 
Further, it is assumed that the tuples are i.i.d.  %
and the joint distribution $P$ factorizes as {\small{$\prod_i P(Z_i)P(T_i\vert Z_i)P(Y_i|Z_i,T_i)$}}.

\textbf{OPE in CBs}
Given data sampled from $P$, the goal in OPE is to evaluate the expected outcome $\E_Q[Y]$ under a distribution $Q$ induced by following a new policy in the \textit{same} environment \citep{dudik2014doubly, thomas2016data}. Importantly, the difference between the two distributions is assumed to be only due to different policies i.e. $P(T\vert Z){\neq} Q(T\vert Z)$ and the rest of the environment-related factors are the same. That is, only shifts on $T$ are considered as the domain expert intends to implement a different policy, $Q(T\vert Z)$ instead of $P(T\vert Z)$.

\subsection{Markov Decision Process} 
Markov Decision Processes are characterized by the tuple $(\cS, \cA, \cP, r, \gamma)$, where $\cS$ is the state-space, $\cA$ is the action-space, $\cP \triangleq P(\cdot\vert s,a)$ characterizes the dynamics, $r: \cS \times \cA \to \mathbb{R}$ is the reward model and $\gamma\in [0,1)$ is the discount factor. 
We restrict ourselves to finite state, finite action infinite-horizon MDPs in this work. %
The the reward model and transition model are assumed to be fixed. %
Value of a policy $\pi: \cS \to \Delta(\cA)$ is defined as the expected cumulative reward received starting from $s_0$,
$$V^\pi(s_0) = \E_{\pi,P}\left[\sum_{t=0}^\infty \gamma^t r(s_t,a_t)\vert s_0\right].$$

\textbf{OPE in MDPs}
In off-policy evaluation, the goal is to estimate the value of a policy $\pi$ given that we have collected a batch of trajectories obtained using an alternative policy $\mu$. We motivate the need for robustness in the OPE task in Section~\ref{sec:fullmdp}. 
\section{Our Framework}
To evaluate policies, we rely on user input to characterize which covariates indeed shift across domains. We then incorporate this domain knowledge in the form of specifying an uncertainty set which characterizes potential shifts relative to the nominal training distribution. Our goal is then to provide realistic off-policy estimates of policy performance under the potential shifts. 

Let $Z \subseteq \X$ be the subset of variables that shift across domains. %
In this case, we assume that the uncertainty set $\cU^{\text{sub}}_P$ %
contains all distributions resulting from shifts in subcovariates $Z$ which are bounded in some metric $\cD$. 
Formally,

\begin{equation}
\label{eq:int_set}
    \begin{aligned}
    \text{[Subcovariate DRO set]} \\
    \cU^{\text{sub}}_P := \Big\{& \nu(Z) \ll P(Z) \ \text{s.t.}\ \ %
    \infdiv{\nu(Z)}{P(Z)} \leq \delta %
    \Big\}
 \end{aligned}
\end{equation}
A key question is the choice of divergence metric $\cD$ that can allow us to incorporate expert knowledge. 
While it is easy for a domain expert to provide precise information on which covariates may shift across domains, the magnitude of shift may not be precisely clear. We capture this complexity using the $\divcvar$-based uncertainty set. 
Existing literature \citep{duchi2019distributionally,li2021evaluating,subbaswamy2021evaluating} has leveraged $\divcvar$ for specifying magnitude of shifts and compute worse-case loss in case of supervised learning. Specifically, \citet{duchi2019distributionally} builds the uncertainty set in such a way that the training population contains at least $\delta$ proportion of the test population.
\begin{equation}
\label{eq:marginal_set_sub_app}
\begin{aligned}
 \mathcal{U}^\text{sub}_P := \{&Q(Z)P(V\setminus Z\vert Z)\ \text{s.t.} \\ &P(Z) = \alpha' Q(Z) + (1-\alpha') Q'(Z), \alpha \leq \alpha' \leq 1\}
\end{aligned}
\end{equation}
where $Q'(Z)$ is any distribution and $\alpha \in (0,1]$ determines the minimum size of the subpopulation shared between train and test.
Thus, $\mathcal{U}^\text{sub}_P$ constrains the ratio $\frac{Q(Z)}{P(Z)}\leq \frac{1}{\alpha}$ for all values of $Z$. 
Equivalently, the set can be expressed using $\infdivcvar{Q}{P} = \log{\sup_{Z\in \texttt{dom}(Z)}} \frac{Q(Z)}{P(Z)}\leq \delta$ for $\delta=\frac{1}{\alpha}$. 
Using the subcovariate uncertainty set $\cU^{\text{sub}}_P$, the worst-case value $\R(\theta, \cU^{\text{sub}}_P)$ in Eq. (\ref{eq:robust}) can be found using techniques from convex duality~\citet{duchi2019distributionally}. Details of the resulting estimator of $\R(\theta, \cU^{\text{sub}}_P)$ are presented in Appendix~\ref{app:dro_bound}. For the remaining discussion, we will assume that we can solve the worst-case optimization problem resulting from subcovariate shifts and focus on how to leverage this optimizer for OPE.
We now present our main technical results, particularly the framework $\ouralg$ for off-policy evaluation (OPE) under realistic shifts where human input is incorporated in terms of knowledge of (sub)covariate shifts we anticipate in practice.

\begin{figure}[t!]
\centering
    \includegraphics[scale=0.5]{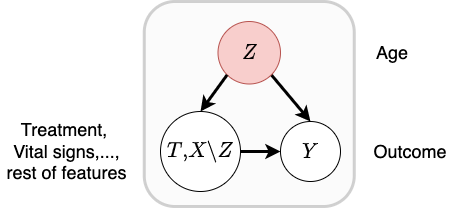}
    \caption{\textbf{Graphical model for shifts in contextual bandits}. Expert input (shaded node $Z$) denotes that $P(Z)$ changes at test time and $P(Y|X,T)$ remains the same.
    For example, treatments $T$ are prescribed based on age $Z$ and have different outcomes $Y$ by age. 
    In practice, only the age distribution may change across environments e.g. across hospitals (as opposed to the joint distribution over age, treatment, and outcome). This motivates a less conservative approach to robustness by focusing on marginal shifts in age.}
    \label{fig:model_cb}
\end{figure}

\subsection{Robust OPE in Contextual Bandits}
\label{sec:offpolicy}
As suggested in Section \ref{sec:prelim_cb}, the bandit setup typically assumes that the joint distribution $P$ changes only due to the change in policy. OPE tasks in CBs focus on evaluating the value of the policy under this assumption. Departing from this assumption, we evaluate the utility of a policy under a \textit{new} environment in which the domain expert anticipates shifts in $Z$. We characterize these by unknown and bounded shifts in $Z$. The following assumption is the CB analogue of the covariate shift assumption in supervised learning.

\begin{assumption}[Expert Input for CBs]
\label{assum:inter_cb}
Suppose the human expert specifies the subset of features $Z\subseteq X$, such that across environments, only $P(Z)$ changes while $P(Y|X,T)$ remains the same.
\end{assumption}

Following this input, the joint distribution at test environment factorizes as $Q(X,T,Y) = Q(Z)Q(T,X\setminus Z\vert Z)P(Y\vert X, T)$. For simplicity, we will group the variables $T,X\setminus Z$ into $T$. The rest of the discussion remains the same and an extra factor of $P(X\setminus Z\vert Z)$ will be suppressed.

Thus, we define the uncertainty set containing distributions $Q$ that
shift in the marginal distribution of $Z$ as,
\begin{equation}\label{eq:unc_cb}
 \begin{aligned}
     \U^{\text{CB}}_P = \{&Q\ll P\ \text{s.t.}\ Q=\nu(Z)Q(T\vert Z)P(Y\vert Z, T), \\ &
     \infdiv{\nu(Z)}{P(Z)}\leq \delta\}
 \end{aligned}
 \end{equation}
The robust OPE problem aims to find the \textit{worst-case} average outcome under $\U^{\text{CB}}_P$ instead of the average,
\begin{equation}
\label{eq:robust_eval}
    \R(\cU^{\text{CB}}_P) = \inf_{Q\in\ \cU^{\text{CB}}_P}\ \E_{(Z, T, Y)\sim Q}[Y],
\end{equation}
where $Q(T\vert Z)$ is the policy to be evaluated and is considered to be known and fixed. Consider each distribution in the set $\cU^{\text{CB}}_P$, $Q(\cdot){=}\nu(Z)Q(T\vert Z)P(Y\vert Z, T)$, which differs from the train distribution in the factors for $Z$ and $T\vert Z$. 
\begin{align}\label{eq:robust_ope_cb}
    &\R(\U^\text{CB}_P) = \inf_{Q\in \U^\text{CB}_P}\ \E_{Z\sim Q(Z)}\E_{P(Y|T,Z)Q(T | Z)}[Y] \\ 
    &\stackrel{(\ref{eq:robust_ope_cb}a)}{=} \inf_{Q \in \U^\text{CB}_P} \E_{Z\sim Q(Z)}\E_{P(Y|T,Z)P(T | Z)}\left[\frac{Q(T|Z)}{P(T|Z)}Y\right] \\
    &\stackrel{(\ref{eq:robust_ope_cb}b)}{=}\inf_{\eta\in \reals} \frac{1}{\delta}\E_{Z \sim P(Z)}\left[(\E\left[W{\times} Y\vert Z\right]-\eta)_+\right] + \eta \nonumberthis
\end{align}
To solve (\ref{eq:robust_eval}) for this $Q$, we first use %
importance sampling to account for the change in $T\vert Z$ due to the known policy $Q(T\vert Z)$, step~(\ref{eq:robust_ope_cb}a). As a result, the set $\U^{\text{CB}}_P$ now consists of shifts on $Z$ alone. Thus, the robust OPE problem reduces to solving $\sup_{Q\in\ \U^{\text{CB}}_P}\ \E_{\V\sim Q}[W{\times} Y]$ where $W$ are the importance sampling weights: $$W(T, Z){=}Q(T\vert Z)/P(T\vert Z)$$ 
Using convex duality arguments~\citet{shapiro2014lectures} for our choice of uncertainty set $\divcvar$, we can obtain  (\ref{eq:robust_ope_cb}b). 
This motivates the full optimization procedure summarized in Algorithm~\ref{alg:cb_ope}. We first compute importance sampling weights $W_i$, and create a re-weighted dataset $\{V_i = (Z_i, W_i \times Y_i)\}_i$. The risk defined in Eq.~\eqref{eq:robust_eval} is approximated by an estimate of its upper bound given in Eq.~\eqref{eq:marginal_smooth_app} when $Z$ are continuous valued (see Appendix \ref{app:dro_bound} for the detailed derivation).

\begin{algorithm}[t!]
\caption{Robust OPE in CBs}
\label{alg:cb_ope}
\begin{algorithmic}
\STATE {\bfseries Input:} Data $\{Z_{i}, T_i, Y_i\}_{i}$, Target policy $Q(T|Z)$, behavior policy $P(T|Z)$, hyperparameters $\delta$, \texttt{L}, \texttt{lr}.
\STATE
\STATE Compute importance weights $\{W_i=\frac{Q(T_i|Z_i)}{P(T_i|Z_i)}\}_i$.
\STATE Create dataset $\{V_i = (Z_{i}, W_i\times Y_i)\}_{i}$.
\STATE Estimate $\widehat{R}(\cU^{\text{CB}}_P)$ (Eq.~(\ref{eq:robust_eval})) with the worst-case risk estimator in Eq. (\ref{eq:marginal_smooth_app}) for the dataset.
\STATE
\STATE{{\bf{return}} $\widehat{R}(\cU^{\text{CB}}_P)$}
\end{algorithmic}
\end{algorithm}

\subsection{Robust OPE in MDPs}
\label{sec:fullmdp}
Off-policy evaluation is critical in sequential decision-making settings, often encountered in human-centered domains such as health. Often such environments are best modeled as a Markov Decision Process (MDP) as introduced in Section~\ref{sec:prelim}. In this case, we have to consider shifts in the transition dynamics across environments which can invalidate OPE methods for MDPs as they often assume stationary dynamics. 

Our goal is to evaluate \textit{robust} value for a given policy $\pi$ to be deployed in a new environment with unknown transition dynamics. Hence, the uncertainty set for each state-action pair is defined over $P(s'|s, a)$, 
denoted by $\mathcal{U}(s,a)$.\footnote{We drop the dependence on the train environment's transition probabilities for conciseness and use $P$ to denote target probabilities instead of $Q$ to avoid confusion with the $Q$-value function in RL.} Specifically, we want to estimate the robust value for $\pi$ starting from $s_0$ as $V^\pi(s_0) = \inf_{P\in\mathcal{U}} \E_{\pi,P}\left[\sum_{t=0}^\infty \gamma^t r(s_t,a_t)\vert s_0\right]$.
 \citet{iyengar2005robust} proves that $V^\pi(\cdot)$ is the solution to the following fixed-point equation (namely, robust Bellman equation) if we assume that uncertainty sets for each state-action pair are constructed independently (known as \textit{SA-rectangularity}),
\begin{equation}
    \label{eq:robust_bellman}
    \begin{aligned}
        V^\pi(s) = r(s,\pi(s)) + \inf_{P\in\mathcal{U}(s,\pi(s))} \gamma \E_{s'\sim P(s'\vert s,\pi(s))}[V^\pi(s')]
    \end{aligned}
\end{equation}
 \begin{figure}[t!]
\centering
        \includegraphics[width=0.4\textwidth]{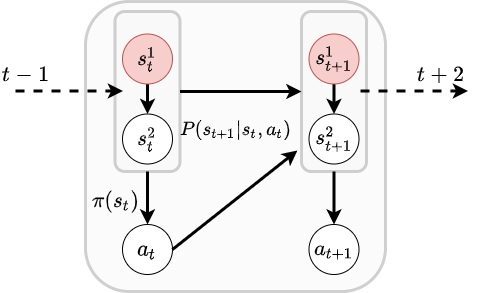}
    \caption{\textbf{Graphical model representing a Markov Decision Process.} Expert input (shaded nodes in red) denotes the variables whose distribution changes across environments. Here, only $P(s_{t+1}^1| s_t,a_t)$ changes. Rewards $r_t$ are not shown for simplicity. Each $r_t$ has directed edges from $s_t$ and $a_t$.
    }
    \label{fig:sel_mdp}
 \end{figure}

Intuitively, SA-rectangularity implies that the uncertainty sets are constructed across time steps independently. This property yields a tractable method to compute the value function estimates using dynamic programming \citep{iyengar2005robust}. We state the assumption formally in Appendix \ref{app:assum_mdp} for completeness.

Eq. (\ref{eq:robust_bellman}) can be solved iteratively by dynamic programming~\citep{sutton2018reinforcement}. Given the value function at any iteration, we additionally have to solve 
the minimization problem over $\mathcal{U}(s,a)$.
Thus, the robust OPE problem in MDPs reduces to solving multiple DRO subproblems with a chosen uncertainty set.

\paragraph{Applying $\ouralg$ to OPE in MDPs.}
Past work has only considered uncertainty sets for the joint distribution \citep{iyengar2005robust,tamar2014scaling,petrik2019beyond,zhou2021tabular}.
In contrast, we consider sets based on the shifts on parts of the state space of the MDP i.e. leveraging human input. Figure \ref{fig:sel_mdp} pictorially represents the probabilistic model for the MDP denoting the shifting state features. The following assumption is the RL analogue of the subcovariate shifts in supervised learning.

For any time step $t>0$ in the MDP, consider a partitioning of the state feature vector $s_t$ into two feature sets, $(s^{1}_t,s^{2}_t)$. The factors of the joint distribution in any environment are given by:
$$P(s_{t+1},r_{t+1}|s_t,a_t) = P(s^{1}_{t+1}\vert s_t, a_t)P(s^{2}_{t+1}\vert s_t, a_t, s^{1}_{t+1})P(r_{t+1}\vert s_t, a_t, s_{t+1})$$

\begin{assumption}[Expert Input for MDPs]
\label{assum:inter_mdp}
Suppose the human expert specifies the following,
\begin{enumerate}[label=(\alph*)]
    \item $P(s^{1}_{t+1}\vert s_t, a_t)$ can shift across environments independently of state-action pairs at any other time step, while
    \item $P(s^{2}_{t+1}\vert s_t, a_t, s^{1}_{t+1})$ and $P(r_{t+1}\vert s_t, a_t, s_{t+1})$ are the same as that in the train environment.
\end{enumerate}
\end{assumption}

\textbf{Remark}. Assumption \ref{assum:inter_mdp} implies that the MDP satisfies SA-rectangularity. For any state-action pair $(s_t,a_t)$ at time step $t$, the shifts that result in the uncertainty set $\U^{\text{MDP}}(s_t,a_t)$ are independent of the state-action pairs at other time steps. This implies that $\U^{\text{MDP}}(s_t,a_t)$ is determined independently of the uncertainty sets at other time steps. Thus, the collection $\U^{\text{MDP}}$ is all possible combinations of sets $\U^{\text{MDP}}(s_t,a_t)$. A scenario where SA-rectangularity does not hold is when uncertainty sets are constructed adaptively based on previous state-action pairs, which we do not address.

Thus, the uncertainty set needs to be defined only for $P(s^{1}_{t+1}\vert s_t, a_t)$, denoted by $\U^{\text{MDP}}(s,a)$:
\begin{align*}
     \U^{\text{MDP}}(s,a) {:=} \Big\{
     &P(\cdot\vert s,a)\ll P_0(\cdot\vert s,a)\quad \text{s.t.}\\
     &\forall s', P(s'\vert s,a)=\nu(s'^{1}\vert s, a)P_0(s'^{2}\vert s, a, s'^{1}), \\ &\infdiv{\nu(s'^{1}\vert s, a)}{P_0(s'^{1}\vert s, a)}\leq \delta\Big\}
\end{align*}
where $P_0(\cdot)$ denotes the distribution of the train environment.
With $\U^{\text{MDP}}$, the DRO subproblem in (\ref{eq:robust_bellman}) reduces to,
\begin{align*}
    &\inf_{P\in\mathcal{U}^{\text{MDP}}(s,\pi(s))} \E_P[V^\pi(s')] \\
    &= \inf_{P\in\mathcal{U}^{\text{MDP}}(s,\pi(s))} \E_{P(s'^{1}\vert s, \pi(s))} \left[\E_{P_0(s'^{2}\vert s, \pi(s), s'^{1})}\left[V^\pi(s')\right]\right]
\end{align*}
We estimate the inner expectation with Monte-Carlo averaging on batch data, followed by solving the DRO problem as done in bandits (\ref{eq:robust_ope_cb}). Here, we use the maximum likelihood estimate of the transition model $P_0$ as we do not have access to the true model. Steps for estimation are outlined in %
Algorithm \ref{alg:cb_mdp}. 

\begin{algorithm}[t!]
\caption{Robust OPE in MDPs}
\label{alg:cb_mdp}
\begin{algorithmic}
\STATE {\bfseries Input:} Trajectories $\{(s,a,s',r)\}$ sampled using policy $\mu$ and transition model $P_0$, Target policy $\pi$, Discount factor $\gamma$, Robustness level $\delta$.
\STATE
\STATE Learn transition models with sample averages across observed trajectories $\widehat{P_0}(s'^1\vert s,a)= \frac{\text{Count}\{(s,a,(s'^1,*),*)\}}{\text{Count}\{(s,a,(*,*),*)\}}$ and $\widehat{P_0}(s'^2\vert s,a,s'^1)= \frac{\text{Count}\{(s,a,(s'^1,s'^2),*)\}}{\text{Count}\{(s,a,(s'^1,*),*)\}}$, where the wildcard $(s,a,*,*)$ denotes transitions from trajectories that match the state-action pair $s,a$.
\STATE
\STATE Learn reward model with sample averages across observed trajectories $\widehat{r}(s,a) = \frac{\sum_{(\_,\_,\_,r)\in \{(s,a,*,*)\}} r}{\text{Count}\{(s,a,*,*)\}}$.
\STATE
\STATE Initialize $V^\pi(s)=0$, for all $s\in\cS$.
\REPEAT
\FOR{$s\in\cS$}
\STATE Update $V^\pi(s)$ using Eq. (\ref{eq:robust_mdp_app}) in Appendix \ref{app:algmdp} with $\widehat{P_0}$.
\ENDFOR
\UNTIL{$V^\pi$ converges}
\STATE
\STATE{{\bf{return}} $V^\pi$}
\end{algorithmic}
\end{algorithm}

Given enough samples of next state for each state-action pair, the robust value can still be estimated accurately. Suppose $V^\pi_{\mathcal{U}_{P_0}}$ be the robust value corresponding to the true transition model $P_0$ which we want to show is close to the robust value corresponding to the \textit{estimated} transition model $\smash{\widehat{P}_0}$ denoted by $\smash{V^\pi_{\mathcal{U}_{\widehat{P}_0}}}$. To show this, we will restrict to uncertainty sets in Eq. (\ref{eq:int_set}) defined by KL-divergence $\infdiv{Q}{P} = \sum_z Q(z)\log(\frac{Q(z)}{P(z)})$ being smaller than $\delta$.
\begin{theorem}[Robust OPE Estimation error]
\label{thm:error_ope}
Given at least $n$ samples from $P_0(\cdot\vert s,a)$ for all $s,a$, assuming that the rewards are bounded $r\in[0,r_\text{max}]$, and that $\mathcal{U}^{\text{MDP}}$ are defined by KL-divergence, then with probability at least $1-\alpha$,
{\small{
\begin{align*}
    \|V^\pi_{\mathcal{U}_{P_0}} - V^\pi_{\mathcal{U}_{\widehat{P}_0}}\|_\infty \leq O\left(\frac{\gamma r_\text{max}|\mathcal{S}|}{(1-\gamma)^2} \sqrt{\frac{1}{n}\log\left(\frac{4|\mathcal{S}\times\mathcal{A}\times\mathcal{S}|}{\alpha}\right)}\right)
\end{align*}
}}
\end{theorem}
Thus, the error in evaluating the robust value with the estimated transition model as compared to the true model converges at the rate $\tilde{O}\left(\frac{\gamma|\mathcal{S}|}{\sqrt{n}(1-\gamma)^2}\right)$, ignoring logarithmic factors. The dependence on $1/\sqrt{n}$ matches that for the non-robust case but is sub-optimal in $|\mathcal{S}|$ \citep{li2020breaking}. The proof is included in Appendix \ref{app:error_ope} in which we show an analogue of the \textit{simulation lemma}~\citep{kearns2002near} for robust MDPs. In terms of time complexity, each iteration of dynamic programming with DRO can be computed in $O(|\mathcal{A}| |\mathcal{S}|^2 \text{log}|\mathcal{S}|)$ which is only a $\text{log}|\mathcal{S}|$ factor more than the non-robust solution.

\begin{figure*}[htbp!]
\centering
\begin{subfigure}[b]{0.45\textwidth}
  \centering
  \includegraphics[width=0.8\textwidth]{{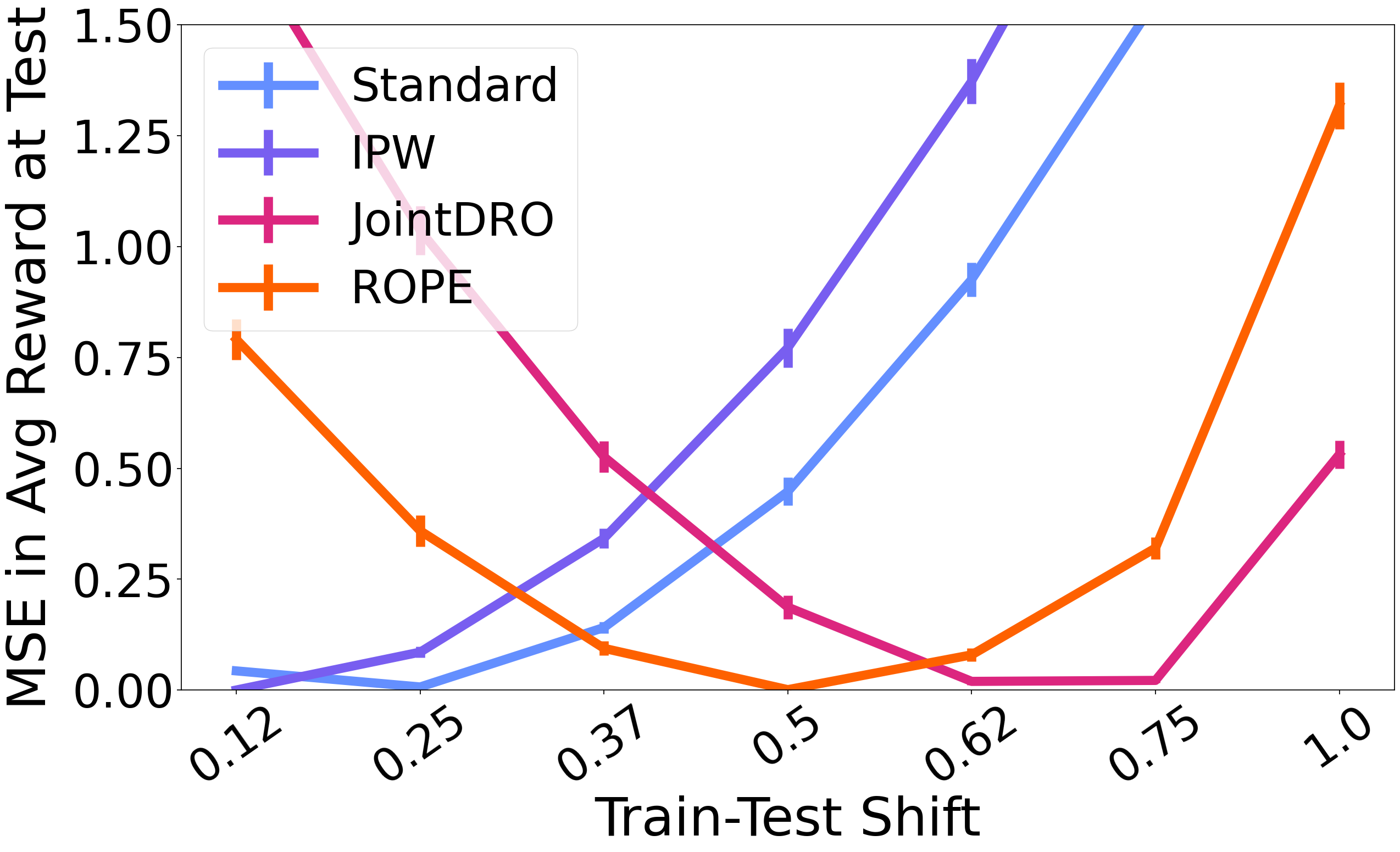}}
  \caption{}
  \label{fig:synth_cb}
\end{subfigure}%
\begin{subfigure}[b]{0.45\textwidth}
  \centering
\includegraphics[width=0.8\textwidth]{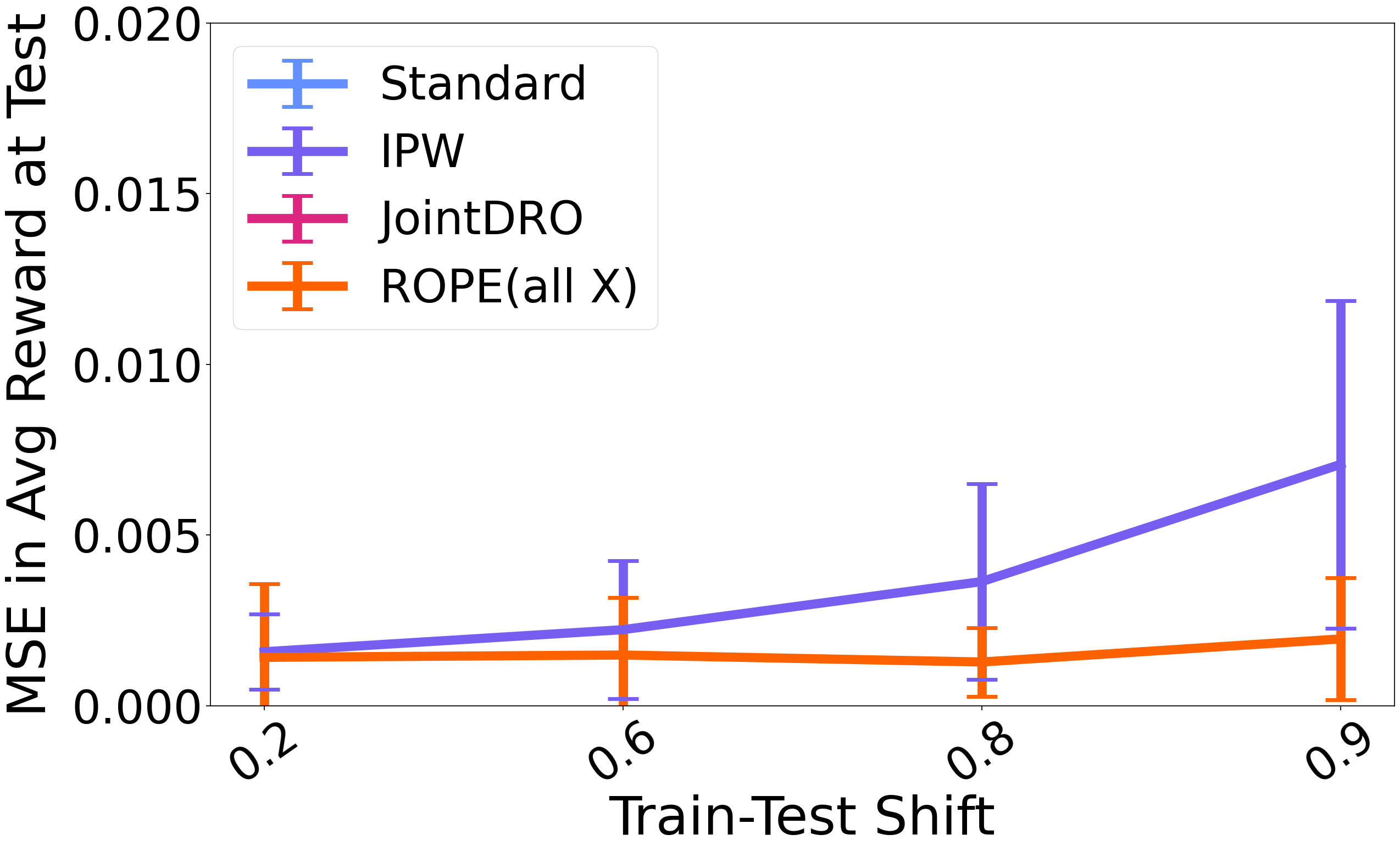}
  \caption{}
  \label{fig:wf_cb}
\end{subfigure}
\caption{(a) \textbf{CB, Synthetic}. MSE in value estimates for test sets (y-axis) with varying levels of shift in $Z_1$ (x-axis). $\ouralg$ performs well for moderate shifts. (b) \textbf{CB, Warfarin}. MSE in value estimates for test sets with shift in the race distribution. $\ouralg$ achieves the right level of conservatism to match the value at test. Curves for $\stdalg$ and $\divalg$ are \textbf{not visible} as they have high error (around 0.1 and 0.7, respectively) and lie outside the plotted y-axis range. 
Robustness level in Eq. (\ref{eq:unc_cb}) is set to $\delta=0.8$. Error bars are computed over $5$ random initializations.}
\label{fig:results}
\end{figure*}

\section{Experimental Evaluation}\label{sec:expts}
In this section, we study the robustness-utility trade-offs achieved by the proposed shifts and compare these with the existing approaches\footnote{Code for replicating the experiments is at \url{https://github.com/AI4LIFE-GROUP/rise-against-distribution-shift}}. More specifically, we study whether optimizing for marginal shifts with $\ouralg$ improves model performance under those shifts in comparison with optimizing for broader classes of shifts using existing methods.
We start with subcovariate shifts in bandit settings as described in Sec. \ref{sec:exp_bandit}. We specifically consider the partial feedback problem in the contextual bandits setup since we only get to the observe the feedback (outcomes) for the actions taken in the collected data. 
Thus, we tackle robustness under partial feedback.
We show that our method provides more faithful estimates for efficacy of a drug dosing policy under patient population shifts. In Sec. \ref{sec:exp_fullmdp}, we consider sequential decision problems under shifts in environment dynamics for MDPs. Within a simulated Sepsis environment, we show that the proposed sets provide significantly less conservative value estimates for a treatment policy.
\subsection{Robust OPE in CBs}
\label{sec:exp_bandit}
We present results on a synthetic and a real world dataset.

\textbf{Synthetic}. We generate data with two features $Z{:=}(Z_1, Z_2)$, binary treatment $T$, and continuous outcome $Y$. Additional details on how data is simulated are deferred to Appendix \ref{app:data_cb}. 
Here we assume changes the marginal distribution of $Z_1$ for evaluation. We simulate $n{=}2000$ samples in the train environment following a logistic policy and use it to estimate the robust value of a different policy in shifted environments. 

Baselines include:
$\stdalg$ returns the average value assuming no shift.
Inverse Probability Weighting \texttt{(IPW)} only corrects for shift in policy using importance sampling. $\divalg$ accounts for shifts in all variables $V$, as done in past work \citep{si2020distributional}.

 In Figure \ref{fig:synth_cb}, we plot the MSE between the estimated and the true policy value, evaluated using $20000$ samples from the test environment. We observe that when the test environment is close to the train one, not accounting for the shift ({$\stdalg$}) performs well. But, as the shift increases, our approach ($\ouralg$) does better. With large shifts, larger uncertainty sets are required. For large shifts, {$\divalg$} does better than the other methods. In summary, $\ouralg$ performs well when the shift is significant but not too large. This highlights the importance of choosing the desired robustness level appropriately which is a challenging open problem for DRO methods.

\begin{table*}%
\centering
\begin{subtable}[t]{0.6\linewidth}
\centering
\begin{tabular}{p{0.7cm}c|cccc}
\toprule
      &    $\stdalg$ & \multicolumn{2}{c}{$\divalg$ } & \multicolumn{2}{c}{$\ouralg$} \\
      $\delta$ &          - &          0.4 &          0.8 &          0.4 &          0.8 \\
\midrule
mean & -1136.43 & -1448.16 & -1221.78 & \textbf{-1416.39} & \textbf{-1190.57} \\
std. &     6.22 &     6.32 &     5.28 &     6.70 &     6.33 \\
median & -1136.64 & -1449.91 & -1222.76 & -1417.12 & -1190.92 \\
\bottomrule
\end{tabular}
\caption{Cliffwalking domain}
\label{tab:cliff_dp}
\end{subtable}\hfill
\begin{subtable}[t]{0.4\linewidth}
\centering
\begin{tabular}{p{0.7cm}c|cccc}
\toprule
    &   $\stdalg$    &  $\divalg$ & $\ouralg$\\
  $\delta$  &   -       &   0.8     &    0.8   \\
\midrule
mean & -0.037 & -0.939 &   \textbf{-0.662} \\
std. &  0.008 &  0.006 &    0.148 \\
median & -0.039 & -0.941 &   -0.705 \\
\bottomrule
\end{tabular}
\caption{Sepsis domain}
\label{tab:sepsis_dp}
\end{subtable}
\caption{\textbf{Robust OPE in MDP.} Estimated value ($10$ random runs) with standard (that is, non-robust) and robust dynamic programming. \ouralg~provides less conservative value than $\divalg$ meaning it has a smaller decrease in value from $\stdalg$.}
\label{tab:mdp_tables}
\end{table*}

\textbf{Warfarin Dosing Policy}. Warfarin is an oral anticoagulant drug. Optimal dosage to assign to a patient while initiating Warfarin therapy has been a subject of multiple clinical trials \citep{heneghan2010optimal}. Using the public PharmGKB dataset~\citep{international2009estimation} of $5528$ patients, \citet{bastani2015online} learn contextual bandit policies that adapt doses based on patient covariates like demographics and clinical information. \footnote{A preprocessed version of the Warfarin dataset was downloaded from \url{https://github.com/khashayarkhv/contextual-bandits/blob/master/datasets/warfarin.csv}.} Reward, either 0 or 1, is defined as a policy's accuracy in making the correct dosing decision.
However, the value of the policies is suspect when applied to patient populations different from the development cohort.
Thus, we estimate robust value of a policy under shifts in race distribution. 
Note that the ground truth optimal dose for each patient is available in the data which enables evaluating different policies.
We learn a dosing policy with linear regression on held-out data and estimate it on test data with shifted race distributions. Specifically, we subsample (without replacement) fewer patients with a recorded race into our analysis set. The policy has lower performance on patients with \texttt{Unknown} race. Thus, the value of the policy decreases with increasing shift as the relative proportion of this group increases. To estimate this value correctly, the robust method must consider the right level of conservativeness.
Figure \ref{fig:wf_cb} shows that MSE between the estimated and true average reward is lower for $\ouralg$ than for $\divalg$ and \texttt{IPW}, as it constructs the sets for marginal shifts alone. %

\begin{figure}[htbp!]
\centering
\includegraphics[width=\linewidth]{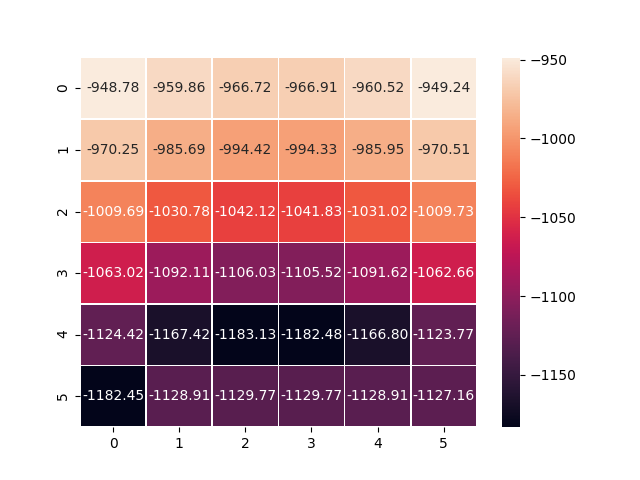}
\caption{\textbf{Illustration of the Cliffwalking domain.} Plot shows (part of) the value function estimated using robust Bellman equation with \ouralg. Start position is (5,0), goal is (5,5), and cliff corresponds to the row (5,0) to (5,5). Agent slips downward by 1 cell with probability 0.1 when taking actions in any of the columns except first and last columns. Results in Table \ref{tab:cliff_dp} report the value at the start position (5,0), which is -1182.45 here, averaged over 10 random initializations of the domain.}
\label{fig:cliff}
\end{figure}

\subsection{Robust OPE in MDPs}
\label{sec:exp_fullmdp}
We present results on two simulated RL domains.

\textbf{Cliffwalking Domain}. We consider a $6\times6$ gridworld which a agent navigates from a start to a goal position avoiding a cliff \citep[][Ex. 6.6]{sutton2018reinforcement}. Please refer to Appendix \ref{app:data_cliff} for more details. An illustration of the gridworld is provided in Figure \ref{fig:cliff}. 
With a constant shift probability, the agent slips towards the cliff instead of taking the prescribed action. The slip probability varies across environments, %
changing the transition dynamics and necessitating robustness in policy evaluation.
We evaluate the value estimate for an agent following uniform random policy using dynamic programming with the standard Bellman equation ($\stdalg$) or the robust one in (\ref{eq:robust_bellman}) ($\divalg,\ouralg$). %
To simulate subcovariate shifts, we duplicate the state features $s^1,s^2$ such that $s^1$ follows the agent's actions while $s^2$ is random noise.
Since agent's actions affect only $s^1$, $\ouralg$ correctly constructs sets on $P(s^1\vert s,a)$. In contrast, $\divalg$ ignores this structure and constructs sets on both the features, $P(s^1,s^2\vert s,a)$. This is the same setting as considered in past work \citep{zhou2021tabular} which used the KL divergence to define uncertainty sets. Table \ref{tab:cliff_dp} reports the value estimates for the start position. We observe that for a high level of desired robustness $\delta=0.4$, $\divalg$ decreases the value by 27.4\% (from $-1136$ to $-1448$) while $\ouralg$ only decreases it by 24.6\% (to $-1416$). 
This validates that both DRO methods have the expected behavior in the MDP. %

\textbf{Sepsis Treatment Evaluation.}
Sepsis simulator~\citep{oberst19counterfactual} is a domain with more involved transition dynamics and has been used to test treatment policies \citep{oberst19counterfactual, namkoong2020off, killian2020counterfactually,futoma2020popcorn}. It has a total of $1440$ states which includes $4$ vital signs (blood pressure, glucose levels, heart rate, and oxygen concentration) and diabetic status. Actions correspond to $3$ treatment combinations
(antibiotics, mechanical vents, and vasopressors). Terminal states, discharge from ICU or death, have rewards $+1$ or $-1$, respectively. %
Glucose levels fluctuate more for diabetics than non-diabetics. Our goal is to evaluate a policy learned using policy iteration, namely RL policy, on a dataset with $20\%$ diabetics. 
We consider a setting where the percentage of diabetics and the fluctuation in their glucose levels varies in the test environment. To interrogate the RL policy for possible deployment, we find its robust value accounting for these shifts. $\divalg$ constructs sets based on the full 1440 states, while $\ouralg$ considers uncertainty only in glucose level dynamics for diabetics and non-diabetics. Thus, $\ouralg$ represents the actual shifts more faithfully compared to $\divalg$. 
We %
check how conservative the OPE estimates from assumptions of joint shifts are relative to leveraging domain knowledge to restrict to subcovariate shifts. %
Table \ref{tab:sepsis_dp} reports the value estimates for the RL policy obtained with standard and robust dynamic programming. We observe that $\divalg$ reports a decrease in value by $21$ times as compared to the value at train environment and $\ouralg$ only reports a decrease in value by $14$ times. Thus, experiments demonstrate the benefit of curating the uncertainty sets using domain knowledge to balance utility and robustness. 

\paragraph{Optimization and system details}
For the CB experiments, we perform gradient descent with Adagrad \citep{duchi11adagrad} implementation in PyTorch \citep{paszke19pytorch} to solve Eq. \ref{eq:marginal_smooth_app}. In Synthetic CB dataset, we use learning rate $0.5$ for $\ouralg$ and $0.1$ for $\divalg$, Lipschitz constant $1$, and number of steps $100$. In Warfarin dataset, we use learning rate $0.1$ for both $\ouralg$ and $\divalg$, Lipschitz constant $0.01$, and number of steps $200$. 
While solving the dynamic program iteratively in the MDP experiments, the convergence condition for $V^\pi$ is taken as maximum change in $V^\pi(\cdot) < 10^{-2}$ after one repetition across all states. The minimization over $\eta$ while solving Eq. (\ref{eq:robust_marginal}) is performed using Brent's method, implemented in the package \texttt{scipy} \citep{scipy}.
For the MDP experiments, the only hyperparameter is $\delta$ which is set to $0.8$ throughout or to $0.4$ in Table \ref{tab:cliff_dp}.
Experiments were run on a compute cluster using a single node with a 2.50 GHz Intel processor, 2 GPUs with 6 GB memory each, and 256 GB system memory. None of the datasets include personally identifiable information.
\section{Conclusions and Future Work}
\label{sec:conc}
In this work we are focused on leveraging human expertise to provide value estimates of ML policies under realistic distribution shifts. 
Here, we propose to represent domain knowledge via uncertainty sets over sub-covariate shifts. %
We argue that this enables representing more realistic shifts and lead to less conservative solutions.
We then provide novel estimators for robust OPE in contextual bandits and MDPs leveraging the distributionally robust optimization framework.
Future directions include expanding human input to %
 address shifts in conditional distributions for OPE (e.g. see \citep{subbaswamy2021evaluating} for supervised learning). 
 Finally, applying the robust OPE method to continuous state-action spaces with function approximators (e.g. \citep{tamar2014scaling}) is an interesting direction of future work. %
We hope that the perspective of %
leveraging human input to define uncertainty sets for robustness in off-policy evaluation opens up more possibilities to tackle over-conservatism of robust learning as well as the challenging problem of model selection in off-policy evaluation.

\paragraph{Broader implications}
Although it is challenging to foresee the impact of using robust methods in real-world applications, one negative is that from over-reliance on results without adequate scrutiny of the assumptions like causal knowledge and uncertainty about future deployment settings. Adversaries can manipulate the deployment environments such that the estimated robust values of policies have high errors. The design of uncertainty sets that account for such adversarial changes should be informed and validated by domain experts. 
\begin{acks}
The authors would like to thank the anonymous reviewers for their feedback and all the funding agencies listed below for supporting this work. This work is supported in part by the Center for Research on Computation and Society (CRCS) at Harvard University, NSF awards \#IIS-2008461, \#IIS-2040989, and \#1922658, and research awards from Amazon, Harvard Data Science Institute, Bayer, and Google. The authors would also like to thank Rediet Abebe, Sara Kingsley, Elita Lobo, Aviva Prins, and Milind Tambe for early feedback on this work. Lastly, HL would like to thank Sujatha and Mohan Lakkaraju for their continued support and encouragement. The views expressed here are those of the authors and do not reflect the official policy or position of the funding agencies.
\end{acks}

\bibliographystyle{ACM-Reference-Format}
\balance
\bibliography{robust}


\begin{thebibliography}{68}


\ifx \showCODEN    \undefined \def \showCODEN     #1{\unskip}     \fi
\ifx \showDOI      \undefined \def \showDOI       #1{#1}\fi
\ifx \showISBNx    \undefined \def \showISBNx     #1{\unskip}     \fi
\ifx \showISBNxiii \undefined \def \showISBNxiii  #1{\unskip}     \fi
\ifx \showISSN     \undefined \def \showISSN      #1{\unskip}     \fi
\ifx \showLCCN     \undefined \def \showLCCN      #1{\unskip}     \fi
\ifx \shownote     \undefined \def \shownote      #1{#1}          \fi
\ifx \showarticletitle \undefined \def \showarticletitle #1{#1}   \fi
\ifx \showURL      \undefined \def \showURL       {\relax}        \fi
\providecommand\bibfield[2]{#2}
\providecommand\bibinfo[2]{#2}
\providecommand\natexlab[1]{#1}
\providecommand\showeprint[2][]{arXiv:#2}

\bibitem[Bastani and Bayati(2015)]%
        {bastani2015online}
\bibfield{author}{\bibinfo{person}{Hamsa Bastani} {and} \bibinfo{person}{Mohsen
  Bayati}.} \bibinfo{year}{2015}\natexlab{}.
\newblock \showarticletitle{Online decision-making with high-dimensional
  covariates}.
\newblock \bibinfo{journal}{\emph{Forthcoming in Operations Research}}
  (\bibinfo{year}{2015}).
\newblock


\bibitem[Ben-Tal et~al\mbox{.}(2013)]%
        {ben2013robust}
\bibfield{author}{\bibinfo{person}{Aharon Ben-Tal}, \bibinfo{person}{Dick
  Den~Hertog}, \bibinfo{person}{Anja De~Waegenaere}, \bibinfo{person}{Bertrand
  Melenberg}, {and} \bibinfo{person}{Gijs Rennen}.}
  \bibinfo{year}{2013}\natexlab{}.
\newblock \showarticletitle{Robust solutions of optimization problems affected
  by uncertain probabilities}.
\newblock \bibinfo{journal}{\emph{Management Science}} \bibinfo{volume}{59},
  \bibinfo{number}{2} (\bibinfo{year}{2013}), \bibinfo{pages}{341--357}.
\newblock


\bibitem[Bertsimas et~al\mbox{.}(2018)]%
        {bertsimas2018data}
\bibfield{author}{\bibinfo{person}{Dimitris Bertsimas}, \bibinfo{person}{Vishal
  Gupta}, {and} \bibinfo{person}{Nathan Kallus}.}
  \bibinfo{year}{2018}\natexlab{}.
\newblock \showarticletitle{Data-driven robust optimization}.
\newblock \bibinfo{journal}{\emph{Mathematical Programming}}
  \bibinfo{volume}{167}, \bibinfo{number}{2} (\bibinfo{year}{2018}),
  \bibinfo{pages}{235--292}.
\newblock


\bibitem[Blanchet and Murthy(2019)]%
        {blanchet2019quantifying}
\bibfield{author}{\bibinfo{person}{Jose Blanchet} {and}
  \bibinfo{person}{Karthyek Murthy}.} \bibinfo{year}{2019}\natexlab{}.
\newblock \showarticletitle{Quantifying distributional model risk via optimal
  transport}.
\newblock \bibinfo{journal}{\emph{Mathematics of Operations Research}}
  \bibinfo{volume}{44}, \bibinfo{number}{2} (\bibinfo{year}{2019}),
  \bibinfo{pages}{565--600}.
\newblock


\bibitem[Canonne(2022)]%
        {canonne2022note}
\bibfield{author}{\bibinfo{person}{Clément~L. Canonne}.}
  \bibinfo{year}{2022}\natexlab{}.
\newblock \bibinfo{title}{A short note on an inequality between KL and TV}.
\newblock
\newblock
\urldef\tempurl%
\url{https://doi.org/10.48550/ARXIV.2202.07198}
\showDOI{\tempurl}


\bibitem[Christiansen et~al\mbox{.}(2020)]%
        {christiansen2020causal}
\bibfield{author}{\bibinfo{person}{Rune Christiansen}, \bibinfo{person}{Niklas
  Pfister}, \bibinfo{person}{Martin~Emil Jakobsen}, \bibinfo{person}{Nicola
  Gnecco}, {and} \bibinfo{person}{Jonas Peters}.}
  \bibinfo{year}{2020}\natexlab{}.
\newblock \showarticletitle{A causal framework for distribution
  generalization}.
\newblock \bibinfo{journal}{\emph{arXiv e-prints}} (\bibinfo{year}{2020}),
  \bibinfo{pages}{arXiv--2006}.
\newblock


\bibitem[Consortium(2009)]%
        {international2009estimation}
\bibfield{author}{\bibinfo{person}{International Warfarin~Pharmacogenetics
  Consortium}.} \bibinfo{year}{2009}\natexlab{}.
\newblock \showarticletitle{Estimation of the warfarin dose with clinical and
  pharmacogenetic data}.
\newblock \bibinfo{journal}{\emph{New England Journal of Medicine}}
  \bibinfo{volume}{360}, \bibinfo{number}{8} (\bibinfo{year}{2009}),
  \bibinfo{pages}{753--764}.
\newblock


\bibitem[Ding et~al\mbox{.}(2021)]%
        {ding2021retiring}
\bibfield{author}{\bibinfo{person}{Frances Ding}, \bibinfo{person}{Moritz
  Hardt}, \bibinfo{person}{John Miller}, {and} \bibinfo{person}{Ludwig
  Schmidt}.} \bibinfo{year}{2021}\natexlab{}.
\newblock \showarticletitle{Retiring Adult: New Datasets for Fair Machine
  Learning}. In \bibinfo{booktitle}{\emph{Advances in Neural Information
  Processing Systems}}, \bibfield{editor}{\bibinfo{person}{A.~Beygelzimer},
  \bibinfo{person}{Y.~Dauphin}, \bibinfo{person}{P.~Liang}, {and}
  \bibinfo{person}{J.~Wortman Vaughan}} (Eds.).
\newblock
\urldef\tempurl%
\url{https://openreview.net/forum?id=bYi_2708mKK}
\showURL{%
\tempurl}


\bibitem[Duchi et~al\mbox{.}(2011)]%
        {duchi11adagrad}
\bibfield{author}{\bibinfo{person}{John Duchi}, \bibinfo{person}{Elad Hazan},
  {and} \bibinfo{person}{Yoram Singer}.} \bibinfo{year}{2011}\natexlab{}.
\newblock \showarticletitle{Adaptive Subgradient Methods for Online Learning
  and Stochastic Optimization}.
\newblock \bibinfo{journal}{\emph{Journal of Machine Learning Research}}
  \bibinfo{volume}{12}, \bibinfo{number}{61} (\bibinfo{year}{2011}),
  \bibinfo{pages}{2121--2159}.
\newblock
\urldef\tempurl%
\url{http://jmlr.org/papers/v12/duchi11a.html}
\showURL{%
\tempurl}


\bibitem[Duchi and Namkoong(2018)]%
        {duchi2018learning}
\bibfield{author}{\bibinfo{person}{John Duchi} {and} \bibinfo{person}{Hongseok
  Namkoong}.} \bibinfo{year}{2018}\natexlab{}.
\newblock \showarticletitle{Learning models with uniform performance via
  distributionally robust optimization}.
\newblock \bibinfo{journal}{\emph{arXiv preprint arXiv:1810.08750}}
  (\bibinfo{year}{2018}).
\newblock


\bibitem[Duchi et~al\mbox{.}(2019)]%
        {duchi2019distributionally}
\bibfield{author}{\bibinfo{person}{John~C Duchi}, \bibinfo{person}{Tatsunori
  Hashimoto}, {and} \bibinfo{person}{Hongseok Namkoong}.}
  \bibinfo{year}{2019}\natexlab{}.
\newblock \showarticletitle{Distributionally robust losses against mixture
  covariate shifts}.
\newblock \bibinfo{journal}{\emph{Under review}} (\bibinfo{year}{2019}).
\newblock


\bibitem[Dud{\'\i}k et~al\mbox{.}(2014)]%
        {dudik2014doubly}
\bibfield{author}{\bibinfo{person}{Miroslav Dud{\'\i}k},
  \bibinfo{person}{Dumitru Erhan}, \bibinfo{person}{John Langford},
  \bibinfo{person}{Lihong Li}, {et~al\mbox{.}}}
  \bibinfo{year}{2014}\natexlab{}.
\newblock \showarticletitle{Doubly robust policy evaluation and optimization}.
\newblock \bibinfo{journal}{\emph{Statist. Sci.}} \bibinfo{volume}{29},
  \bibinfo{number}{4} (\bibinfo{year}{2014}), \bibinfo{pages}{485--511}.
\newblock


\bibitem[Faury et~al\mbox{.}(2020)]%
        {faury2020distributionally}
\bibfield{author}{\bibinfo{person}{Louis Faury}, \bibinfo{person}{Ugo
  Tanielian}, \bibinfo{person}{Elvis Dohmatob}, \bibinfo{person}{Elena
  Smirnova}, {and} \bibinfo{person}{Flavian Vasile}.}
  \bibinfo{year}{2020}\natexlab{}.
\newblock \showarticletitle{Distributionally robust counterfactual risk
  minimization}. In \bibinfo{booktitle}{\emph{Proceedings of the AAAI
  Conference on Artificial Intelligence}}, Vol.~\bibinfo{volume}{34}.
  \bibinfo{pages}{3850--3857}.
\newblock


\bibitem[Futoma et~al\mbox{.}(2020)]%
        {futoma2020popcorn}
\bibfield{author}{\bibinfo{person}{Joseph Futoma}, \bibinfo{person}{Michael~C
  Hughes}, {and} \bibinfo{person}{Finale Doshi-Velez}.}
  \bibinfo{year}{2020}\natexlab{}.
\newblock \showarticletitle{Popcorn: Partially observed prediction constrained
  reinforcement learning}.
\newblock \bibinfo{journal}{\emph{arXiv preprint arXiv:2001.04032}}
  (\bibinfo{year}{2020}).
\newblock


\bibitem[Goodfellow et~al\mbox{.}(2014)]%
        {goodfellow2014explaining}
\bibfield{author}{\bibinfo{person}{Ian~J Goodfellow}, \bibinfo{person}{Jonathon
  Shlens}, {and} \bibinfo{person}{Christian Szegedy}.}
  \bibinfo{year}{2014}\natexlab{}.
\newblock \showarticletitle{Explaining and harnessing adversarial examples}.
\newblock \bibinfo{journal}{\emph{arXiv preprint arXiv:1412.6572}}
  (\bibinfo{year}{2014}).
\newblock


\bibitem[Gottesman et~al\mbox{.}(2018)]%
        {gottesman2018evaluating}
\bibfield{author}{\bibinfo{person}{Omer Gottesman}, \bibinfo{person}{Fredrik
  Johansson}, \bibinfo{person}{Joshua Meier}, \bibinfo{person}{Jack Dent},
  \bibinfo{person}{Donghun Lee}, \bibinfo{person}{Srivatsan Srinivasan},
  \bibinfo{person}{Linying Zhang}, \bibinfo{person}{Yi Ding},
  \bibinfo{person}{David Wihl}, \bibinfo{person}{Xuefeng Peng},
  \bibinfo{person}{Jiayu Yao}, \bibinfo{person}{Isaac Lage},
  \bibinfo{person}{Christopher Mosch}, \bibinfo{person}{Li-wei~H. Lehman},
  \bibinfo{person}{Matthieu Komorowski}, \bibinfo{person}{Matthieu Komorowski},
  \bibinfo{person}{Aldo Faisal}, \bibinfo{person}{Leo~Anthony Celi},
  \bibinfo{person}{David Sontag}, {and} \bibinfo{person}{Finale Doshi-Velez}.}
  \bibinfo{year}{2018}\natexlab{}.
\newblock \bibinfo{title}{Evaluating Reinforcement Learning Algorithms in
  Observational Health Settings}.
\newblock
\newblock
\urldef\tempurl%
\url{https://doi.org/10.48550/ARXIV.1805.12298}
\showDOI{\tempurl}


\bibitem[Hatt et~al\mbox{.}(2021)]%
        {hatt2021generalizing}
\bibfield{author}{\bibinfo{person}{Tobias Hatt}, \bibinfo{person}{Daniel
  Tschernutter}, {and} \bibinfo{person}{Stefan Feuerriegel}.}
  \bibinfo{year}{2021}\natexlab{}.
\newblock \bibinfo{title}{Generalizing Off-Policy Learning under Sample
  Selection Bias}.
\newblock
\newblock
\showeprint[arxiv]{2112.01387}~[stat.ML]


\bibitem[Heneghan et~al\mbox{.}(2010)]%
        {heneghan2010optimal}
\bibfield{author}{\bibinfo{person}{Carl Heneghan}, \bibinfo{person}{Sally
  Tyndel}, \bibinfo{person}{Clare Bankhead}, \bibinfo{person}{Yi Wan},
  \bibinfo{person}{David Keeling}, \bibinfo{person}{Rafael Perera}, {and}
  \bibinfo{person}{Alison Ward}.} \bibinfo{year}{2010}\natexlab{}.
\newblock \showarticletitle{Optimal loading dose for the initiation of
  warfarin: a systematic review}.
\newblock \bibinfo{journal}{\emph{BMC cardiovascular disorders}}
  \bibinfo{volume}{10}, \bibinfo{number}{1} (\bibinfo{year}{2010}),
  \bibinfo{pages}{1--12}.
\newblock


\bibitem[Hoeffding(1994)]%
        {hoeffding1994probability}
\bibfield{author}{\bibinfo{person}{Wassily Hoeffding}.}
  \bibinfo{year}{1994}\natexlab{}.
\newblock \showarticletitle{Probability inequalities for sums of bounded random
  variables}.
\newblock In \bibinfo{booktitle}{\emph{The Collected Works of Wassily
  Hoeffding}}. \bibinfo{publisher}{Springer}, \bibinfo{pages}{409--426}.
\newblock


\bibitem[Hu et~al\mbox{.}(2018)]%
        {hu2018does}
\bibfield{author}{\bibinfo{person}{Weihua Hu}, \bibinfo{person}{Gang Niu},
  \bibinfo{person}{Issei Sato}, {and} \bibinfo{person}{Masashi Sugiyama}.}
  \bibinfo{year}{2018}\natexlab{}.
\newblock \showarticletitle{Does distributionally robust supervised learning
  give robust classifiers?}. In \bibinfo{booktitle}{\emph{International
  Conference on Machine Learning}}. \bibinfo{pages}{2029--2037}.
\newblock


\bibitem[Iyengar(2005)]%
        {iyengar2005robust}
\bibfield{author}{\bibinfo{person}{Garud~N Iyengar}.}
  \bibinfo{year}{2005}\natexlab{}.
\newblock \showarticletitle{Robust dynamic programming}.
\newblock \bibinfo{journal}{\emph{Mathematics of Operations Research}}
  \bibinfo{volume}{30}, \bibinfo{number}{2} (\bibinfo{year}{2005}),
  \bibinfo{pages}{257--280}.
\newblock


\bibitem[Jeong and Namkoong(2020)]%
        {jeong2020robust}
\bibfield{author}{\bibinfo{person}{Sookyo Jeong} {and}
  \bibinfo{person}{Hongseok Namkoong}.} \bibinfo{year}{2020}\natexlab{}.
\newblock \showarticletitle{Robust causal inference under covariate shift via
  worst-case subpopulation treatment effects}. In
  \bibinfo{booktitle}{\emph{Conference on Learning Theory}}. PMLR,
  \bibinfo{pages}{2079--2084}.
\newblock


\bibitem[Jiang(2020)]%
        {jiang2020notes}
\bibfield{author}{\bibinfo{person}{Nan Jiang}.}
  \bibinfo{year}{2020}\natexlab{}.
\newblock \bibinfo{title}{Notes on Tabular Methods}.
\newblock
  \bibinfo{howpublished}{\url{https://nanjiang.cs.illinois.edu/files/cs598/note3.pdf}}.
\newblock
\newblock
\shownote{Accessed: {2021–06-04}}.


\bibitem[Johnson et~al\mbox{.}(2018)]%
        {johnson2018generalizability}
\bibfield{author}{\bibinfo{person}{Alistair E.~W. Johnson},
  \bibinfo{person}{Tom~J. Pollard}, {and} \bibinfo{person}{Tristan Naumann}.}
  \bibinfo{year}{2018}\natexlab{}.
\newblock \bibinfo{title}{Generalizability of predictive models for intensive
  care unit patients}.
\newblock
\newblock
\showeprint[arxiv]{1812.02275}~[cs.LG]


\bibitem[Kearns and Singh(2002)]%
        {kearns2002near}
\bibfield{author}{\bibinfo{person}{Michael Kearns} {and}
  \bibinfo{person}{Satinder Singh}.} \bibinfo{year}{2002}\natexlab{}.
\newblock \showarticletitle{Near-optimal reinforcement learning in polynomial
  time}.
\newblock \bibinfo{journal}{\emph{Machine learning}} \bibinfo{volume}{49},
  \bibinfo{number}{2} (\bibinfo{year}{2002}), \bibinfo{pages}{209--232}.
\newblock


\bibitem[Killian et~al\mbox{.}(2020)]%
        {killian2020counterfactually}
\bibfield{author}{\bibinfo{person}{Taylor~W Killian}, \bibinfo{person}{Marzyeh
  Ghassemi}, {and} \bibinfo{person}{Shalmali Joshi}.}
  \bibinfo{year}{2020}\natexlab{}.
\newblock \showarticletitle{Counterfactually Guided Policy Transfer in Clinical
  Settings}.
\newblock \bibinfo{journal}{\emph{arXiv preprint arXiv:2006.11654}}
  (\bibinfo{year}{2020}).
\newblock


\bibitem[Kulinski et~al\mbox{.}(2020)]%
        {kulinski2020feature}
\bibfield{author}{\bibinfo{person}{Sean Kulinski}, \bibinfo{person}{Saurabh
  Bagchi}, {and} \bibinfo{person}{David~I. Inouye}.}
  \bibinfo{year}{2020}\natexlab{}.
\newblock \showarticletitle{Feature Shift Detection: Localizing Which Features
  Have Shifted via Conditional Distribution Tests}. In
  \bibinfo{booktitle}{\emph{Neural Information Processing Systems (NeurIPS)}}.
\newblock


\bibitem[Laidlaw et~al\mbox{.}(2021)]%
        {laidlaw2021perceptual}
\bibfield{author}{\bibinfo{person}{Cassidy Laidlaw}, \bibinfo{person}{Sahil
  Singla}, {and} \bibinfo{person}{Soheil Feizi}.}
  \bibinfo{year}{2021}\natexlab{}.
\newblock \showarticletitle{Perceptual Adversarial Robustness: Defense Against
  Unseen Threat Models}. In \bibinfo{booktitle}{\emph{ICLR}}.
\newblock


\bibitem[Levin and Peres(2017)]%
        {levin2017markov}
\bibfield{author}{\bibinfo{person}{David~A Levin} {and} \bibinfo{person}{Yuval
  Peres}.} \bibinfo{year}{2017}\natexlab{}.
\newblock \bibinfo{booktitle}{\emph{Markov chains and mixing times}}.
  Vol.~\bibinfo{volume}{107}.
\newblock \bibinfo{publisher}{American Mathematical Soc.}
\newblock


\bibitem[Levine et~al\mbox{.}(2020)]%
        {levine2020offline}
\bibfield{author}{\bibinfo{person}{Sergey Levine}, \bibinfo{person}{Aviral
  Kumar}, \bibinfo{person}{George Tucker}, {and} \bibinfo{person}{Justin Fu}.}
  \bibinfo{year}{2020}\natexlab{}.
\newblock \bibinfo{title}{Offline Reinforcement Learning: Tutorial, Review, and
  Perspectives on Open Problems}.
\newblock
\newblock
\showeprint[arxiv]{2005.01643}~[cs.LG]


\bibitem[Li et~al\mbox{.}(2020)]%
        {li2020breaking}
\bibfield{author}{\bibinfo{person}{Gen Li}, \bibinfo{person}{Yuting Wei},
  \bibinfo{person}{Yuejie Chi}, \bibinfo{person}{Yuantao Gu}, {and}
  \bibinfo{person}{Yuxin Chen}.} \bibinfo{year}{2020}\natexlab{}.
\newblock \showarticletitle{Breaking the Sample Size Barrier in Model-Based
  Reinforcement Learning with a Generative Model}. In
  \bibinfo{booktitle}{\emph{Advances in Neural Information Processing
  Systems}}, \bibfield{editor}{\bibinfo{person}{H.~Larochelle},
  \bibinfo{person}{M.~Ranzato}, \bibinfo{person}{R.~Hadsell},
  \bibinfo{person}{M.~F. Balcan}, {and} \bibinfo{person}{H.~Lin}} (Eds.),
  Vol.~\bibinfo{volume}{33}. \bibinfo{publisher}{Curran Associates, Inc.},
  \bibinfo{pages}{12861--12872}.
\newblock
\urldef\tempurl%
\url{https://proceedings.neurips.cc/paper/2020/file/96ea64f3a1aa2fd00c72faacf0cb8ac9-Paper.pdf}
\showURL{%
\tempurl}


\bibitem[Li et~al\mbox{.}(2018)]%
        {min2018sba}
\bibfield{author}{\bibinfo{person}{Min Li}, \bibinfo{person}{Amy Mickel}, {and}
  \bibinfo{person}{Stanley Taylor}.} \bibinfo{year}{2018}\natexlab{}.
\newblock \showarticletitle{“Should This Loan be Approved or Denied?”: A
  Large Dataset with Class Assignment Guidelines}.
\newblock \bibinfo{journal}{\emph{Journal of Statistics Education}}
  \bibinfo{volume}{26}, \bibinfo{number}{1} (\bibinfo{year}{2018}),
  \bibinfo{pages}{55--66}.
\newblock
\urldef\tempurl%
\url{https://doi.org/10.1080/10691898.2018.1434342}
\showDOI{\tempurl}
\showeprint{https://doi.org/10.1080/10691898.2018.1434342}


\bibitem[Li et~al\mbox{.}(2021)]%
        {li2021evaluating}
\bibfield{author}{\bibinfo{person}{Mike Li}, \bibinfo{person}{Hongseok
  Namkoong}, {and} \bibinfo{person}{Shangzhou Xia}.}
  \bibinfo{year}{2021}\natexlab{}.
\newblock \showarticletitle{Evaluating model performance under worst-case
  subpopulations}. In \bibinfo{booktitle}{\emph{Advances in Neural Information
  Processing Systems}}, \bibfield{editor}{\bibinfo{person}{A.~Beygelzimer},
  \bibinfo{person}{Y.~Dauphin}, \bibinfo{person}{P.~Liang}, {and}
  \bibinfo{person}{J.~Wortman Vaughan}} (Eds.).
\newblock
\urldef\tempurl%
\url{https://openreview.net/forum?id=nehzxAdyJxF}
\showURL{%
\tempurl}


\bibitem[Madry et~al\mbox{.}(2018)]%
        {madry2018towards}
\bibfield{author}{\bibinfo{person}{Aleksander Madry},
  \bibinfo{person}{Aleksandar Makelov}, \bibinfo{person}{Ludwig Schmidt},
  \bibinfo{person}{Dimitris Tsipras}, {and} \bibinfo{person}{Adrian Vladu}.}
  \bibinfo{year}{2018}\natexlab{}.
\newblock \showarticletitle{Towards Deep Learning Models Resistant to
  Adversarial Attacks}. In \bibinfo{booktitle}{\emph{International Conference
  on Learning Representations}}.
\newblock


\bibitem[Magliacane et~al\mbox{.}(2018)]%
        {magliacane2018domain}
\bibfield{author}{\bibinfo{person}{Sara Magliacane}, \bibinfo{person}{Thijs van
  Ommen}, \bibinfo{person}{Tom Claassen}, \bibinfo{person}{Stephan Bongers},
  \bibinfo{person}{Philip Versteeg}, {and} \bibinfo{person}{Joris~M Mooij}.}
  \bibinfo{year}{2018}\natexlab{}.
\newblock \showarticletitle{Domain adaptation by using causal inference to
  predict invariant conditional distributions}. In
  \bibinfo{booktitle}{\emph{Advances in Neural Information Processing
  Systems}}. \bibinfo{pages}{10846--10856}.
\newblock


\bibitem[Maini et~al\mbox{.}(2020)]%
        {maini20union}
\bibfield{author}{\bibinfo{person}{Pratyush Maini}, \bibinfo{person}{Eric
  Wong}, {and} \bibinfo{person}{Zico Kolter}.} \bibinfo{year}{2020}\natexlab{}.
\newblock \showarticletitle{Adversarial Robustness Against the Union of
  Multiple Perturbation Models}. In \bibinfo{booktitle}{\emph{Proceedings of
  the 37th International Conference on Machine Learning}}
  \emph{(\bibinfo{series}{Proceedings of Machine Learning Research},
  Vol.~\bibinfo{volume}{119})}, \bibfield{editor}{\bibinfo{person}{Hal~Daumé
  III} {and} \bibinfo{person}{Aarti Singh}} (Eds.). \bibinfo{publisher}{PMLR},
  \bibinfo{pages}{6640--6650}.
\newblock
\urldef\tempurl%
\url{http://proceedings.mlr.press/v119/maini20a.html}
\showURL{%
\tempurl}


\bibitem[Miller et~al\mbox{.}(2020)]%
        {miller20qamodels}
\bibfield{author}{\bibinfo{person}{John Miller}, \bibinfo{person}{Karl Krauth},
  \bibinfo{person}{Benjamin Recht}, {and} \bibinfo{person}{Ludwig Schmidt}.}
  \bibinfo{year}{2020}\natexlab{}.
\newblock \showarticletitle{The Effect of Natural Distribution Shift on
  Question Answering Models}. In \bibinfo{booktitle}{\emph{Proceedings of the
  37th International Conference on Machine Learning}}
  \emph{(\bibinfo{series}{Proceedings of Machine Learning Research},
  Vol.~\bibinfo{volume}{119})}, \bibfield{editor}{\bibinfo{person}{Hal~Daumé
  III} {and} \bibinfo{person}{Aarti Singh}} (Eds.). \bibinfo{publisher}{PMLR},
  \bibinfo{pages}{6905--6916}.
\newblock
\urldef\tempurl%
\url{http://proceedings.mlr.press/v119/miller20a.html}
\showURL{%
\tempurl}


\bibitem[Mo et~al\mbox{.}(2020)]%
        {mo2020learning}
\bibfield{author}{\bibinfo{person}{Weibin Mo}, \bibinfo{person}{Zhengling Qi},
  {and} \bibinfo{person}{Yufeng Liu}.} \bibinfo{year}{2020}\natexlab{}.
\newblock \showarticletitle{Learning optimal distributionally robust
  individualized treatment rules}.
\newblock \bibinfo{journal}{\emph{J. Amer. Statist. Assoc.}}
  (\bibinfo{year}{2020}), \bibinfo{pages}{1--16}.
\newblock


\bibitem[Namkoong et~al\mbox{.}(2020)]%
        {namkoong2020off}
\bibfield{author}{\bibinfo{person}{Hongseok Namkoong}, \bibinfo{person}{Ramtin
  Keramati}, \bibinfo{person}{Steve Yadlowsky}, {and} \bibinfo{person}{Emma
  Brunskill}.} \bibinfo{year}{2020}\natexlab{}.
\newblock \showarticletitle{Off-policy Policy Evaluation For Sequential
  Decisions Under Unobserved Confounding}. In
  \bibinfo{booktitle}{\emph{Advances in Neural Information Processing
  Systems}}.
\newblock


\bibitem[Nilim and El~Ghaoui(2005)]%
        {nilim2005robust}
\bibfield{author}{\bibinfo{person}{Arnab Nilim} {and} \bibinfo{person}{Laurent
  El~Ghaoui}.} \bibinfo{year}{2005}\natexlab{}.
\newblock \showarticletitle{Robust control of Markov decision processes with
  uncertain transition matrices}.
\newblock \bibinfo{journal}{\emph{Operations Research}} \bibinfo{volume}{53},
  \bibinfo{number}{5} (\bibinfo{year}{2005}), \bibinfo{pages}{780--798}.
\newblock


\bibitem[Oberst and Sontag(2019)]%
        {oberst19counterfactual}
\bibfield{author}{\bibinfo{person}{Michael Oberst} {and} \bibinfo{person}{David
  Sontag}.} \bibinfo{year}{2019}\natexlab{}.
\newblock \showarticletitle{Counterfactual Off-Policy Evaluation with
  {G}umbel-Max Structural Causal Models}. In
  \bibinfo{booktitle}{\emph{Proceedings of the 36th International Conference on
  Machine Learning}} \emph{(\bibinfo{series}{Proceedings of Machine Learning
  Research}, Vol.~\bibinfo{volume}{97})},
  \bibfield{editor}{\bibinfo{person}{Kamalika Chaudhuri} {and}
  \bibinfo{person}{Ruslan Salakhutdinov}} (Eds.). \bibinfo{publisher}{PMLR},
  \bibinfo{pages}{4881--4890}.
\newblock
\urldef\tempurl%
\url{http://proceedings.mlr.press/v97/oberst19a.html}
\showURL{%
\tempurl}


\bibitem[Oberst et~al\mbox{.}(2021)]%
        {oberst2021regularizing}
\bibfield{author}{\bibinfo{person}{Michael Oberst}, \bibinfo{person}{Nikolaj
  Thams}, \bibinfo{person}{Jonas Peters}, {and} \bibinfo{person}{David
  Sontag}.} \bibinfo{year}{2021}\natexlab{}.
\newblock \showarticletitle{Regularizing towards Causal Invariance: Linear
  Models with Proxies}.
\newblock \bibinfo{journal}{\emph{arXiv preprint arXiv:2103.02477}}
  (\bibinfo{year}{2021}).
\newblock


\bibitem[Paszke et~al\mbox{.}(2019)]%
        {paszke19pytorch}
\bibfield{author}{\bibinfo{person}{Adam Paszke}, \bibinfo{person}{Sam Gross},
  \bibinfo{person}{Francisco Massa}, \bibinfo{person}{Adam Lerer},
  \bibinfo{person}{James Bradbury}, \bibinfo{person}{Gregory Chanan},
  \bibinfo{person}{Trevor Killeen}, \bibinfo{person}{Zeming Lin},
  \bibinfo{person}{Natalia Gimelshein}, \bibinfo{person}{Luca Antiga},
  \bibinfo{person}{Alban Desmaison}, \bibinfo{person}{Andreas Kopf},
  \bibinfo{person}{Edward Yang}, \bibinfo{person}{Zachary DeVito},
  \bibinfo{person}{Martin Raison}, \bibinfo{person}{Alykhan Tejani},
  \bibinfo{person}{Sasank Chilamkurthy}, \bibinfo{person}{Benoit Steiner},
  \bibinfo{person}{Lu Fang}, \bibinfo{person}{Junjie Bai}, {and}
  \bibinfo{person}{Soumith Chintala}.} \bibinfo{year}{2019}\natexlab{}.
\newblock \showarticletitle{PyTorch: An Imperative Style, High-Performance Deep
  Learning Library}.
\newblock In \bibinfo{booktitle}{\emph{Advances in Neural Information
  Processing Systems 32}}, \bibfield{editor}{\bibinfo{person}{H.~Wallach},
  \bibinfo{person}{H.~Larochelle}, \bibinfo{person}{A.~Beygelzimer},
  \bibinfo{person}{F.~d\textquotesingle Alch\'{e}-Buc},
  \bibinfo{person}{E.~Fox}, {and} \bibinfo{person}{R.~Garnett}} (Eds.).
  \bibinfo{publisher}{Curran Associates, Inc.}, \bibinfo{pages}{8024--8035}.
\newblock
\urldef\tempurl%
\url{https://proceedings.neurips.cc/paper/2019/file/bdbca288fee7f92f2bfa9f7012727740-Paper.pdf}
\showURL{%
\tempurl}


\bibitem[Peine et~al\mbox{.}(2021)]%
        {peine2021development}
\bibfield{author}{\bibinfo{person}{Arne Peine}, \bibinfo{person}{Ahmed
  Hallawa}, \bibinfo{person}{Johannes Bickenbach}, \bibinfo{person}{Guido
  Dartmann}, \bibinfo{person}{Lejla~Begic Fazlic}, \bibinfo{person}{Anke
  Schmeink}, \bibinfo{person}{Gerd Ascheid}, \bibinfo{person}{Christoph
  Thiemermann}, \bibinfo{person}{Andreas Schuppert}, \bibinfo{person}{Ryan
  Kindle}, {et~al\mbox{.}}} \bibinfo{year}{2021}\natexlab{}.
\newblock \showarticletitle{Development and validation of a reinforcement
  learning algorithm to dynamically optimize mechanical ventilation in critical
  care}.
\newblock \bibinfo{journal}{\emph{NPJ digital medicine}} \bibinfo{volume}{4},
  \bibinfo{number}{1} (\bibinfo{year}{2021}), \bibinfo{pages}{1--12}.
\newblock


\bibitem[Peters et~al\mbox{.}(2016)]%
        {peters2016causal}
\bibfield{author}{\bibinfo{person}{Jonas Peters}, \bibinfo{person}{Peter
  B{\"u}hlmann}, {and} \bibinfo{person}{Nicolai Meinshausen}.}
  \bibinfo{year}{2016}\natexlab{}.
\newblock \showarticletitle{Causal inference by using invariant prediction:
  identification and confidence intervals}.
\newblock \bibinfo{journal}{\emph{Journal of the Royal Statistical Society:
  Series B (Statistical Methodology)}} \bibinfo{volume}{78},
  \bibinfo{number}{5} (\bibinfo{year}{2016}), \bibinfo{pages}{947--1012}.
\newblock


\bibitem[Petrik and Russel(2019)]%
        {petrik2019beyond}
\bibfield{author}{\bibinfo{person}{Marek Petrik} {and}
  \bibinfo{person}{Reazul~Hasan Russel}.} \bibinfo{year}{2019}\natexlab{}.
\newblock \showarticletitle{Beyond confidence regions: Tight bayesian ambiguity
  sets for robust mdps}. In \bibinfo{booktitle}{\emph{Advances in Neural
  Information Processing Systems}}. \bibinfo{pages}{7049--7058}.
\newblock


\bibitem[Pinto et~al\mbox{.}(2017)]%
        {pinto2017adversarial}
\bibfield{author}{\bibinfo{person}{Lerrel Pinto}, \bibinfo{person}{James
  Davidson}, \bibinfo{person}{Rahul Sukthankar}, {and} \bibinfo{person}{Abhinav
  Gupta}.} \bibinfo{year}{2017}\natexlab{}.
\newblock \showarticletitle{Robust Adversarial Reinforcement Learning}. In
  \bibinfo{booktitle}{\emph{Proceedings of the 34th International Conference on
  Machine Learning}} \emph{(\bibinfo{series}{Proceedings of Machine Learning
  Research}, Vol.~\bibinfo{volume}{70})},
  \bibfield{editor}{\bibinfo{person}{Doina Precup} {and}
  \bibinfo{person}{Yee~Whye Teh}} (Eds.). \bibinfo{publisher}{PMLR},
  \bibinfo{pages}{2817--2826}.
\newblock
\urldef\tempurl%
\url{http://proceedings.mlr.press/v70/pinto17a.html}
\showURL{%
\tempurl}


\bibitem[Qi and Liao(2020)]%
        {qi2020robust}
\bibfield{author}{\bibinfo{person}{Zhengling Qi} {and} \bibinfo{person}{Peng
  Liao}.} \bibinfo{year}{2020}\natexlab{}.
\newblock \showarticletitle{Robust Batch Policy Learning in Markov Decision
  Processes}.
\newblock \bibinfo{journal}{\emph{arXiv preprint arXiv:2011.04185}}
  (\bibinfo{year}{2020}).
\newblock


\bibitem[Raghunathan et~al\mbox{.}(2020)]%
        {raghunathan2020understanding}
\bibfield{author}{\bibinfo{person}{Aditi Raghunathan},
  \bibinfo{person}{Sang~Michael Xie}, \bibinfo{person}{Fanny Yang},
  \bibinfo{person}{John Duchi}, {and} \bibinfo{person}{Percy Liang}.}
  \bibinfo{year}{2020}\natexlab{}.
\newblock \showarticletitle{Understanding and Mitigating the Tradeoff between
  Robustness and Accuracy}. In \bibinfo{booktitle}{\emph{Proceedings of the
  37th International Conference on Machine Learning}}
  \emph{(\bibinfo{series}{Proceedings of Machine Learning Research},
  Vol.~\bibinfo{volume}{119})}, \bibfield{editor}{\bibinfo{person}{Hal~Daumé
  III} {and} \bibinfo{person}{Aarti Singh}} (Eds.). \bibinfo{publisher}{PMLR},
  \bibinfo{pages}{7909--7919}.
\newblock
\urldef\tempurl%
\url{https://proceedings.mlr.press/v119/raghunathan20a.html}
\showURL{%
\tempurl}


\bibitem[Rockafellar et~al\mbox{.}(2000)]%
        {rockafellar2000optimization}
\bibfield{author}{\bibinfo{person}{R~Tyrrell Rockafellar},
  \bibinfo{person}{Stanislav Uryasev}, {et~al\mbox{.}}}
  \bibinfo{year}{2000}\natexlab{}.
\newblock \showarticletitle{Optimization of conditional value-at-risk}.
\newblock \bibinfo{journal}{\emph{Journal of risk}}  \bibinfo{volume}{2}
  (\bibinfo{year}{2000}), \bibinfo{pages}{21--42}.
\newblock


\bibitem[Rojas-Carulla et~al\mbox{.}(2018)]%
        {rojas2018invariant}
\bibfield{author}{\bibinfo{person}{Mateo Rojas-Carulla},
  \bibinfo{person}{Bernhard Sch{\"o}lkopf}, \bibinfo{person}{Richard Turner},
  {and} \bibinfo{person}{Jonas Peters}.} \bibinfo{year}{2018}\natexlab{}.
\newblock \showarticletitle{Invariant models for causal transfer learning}.
\newblock \bibinfo{journal}{\emph{The Journal of Machine Learning Research}}
  \bibinfo{volume}{19}, \bibinfo{number}{1} (\bibinfo{year}{2018}),
  \bibinfo{pages}{1309--1342}.
\newblock


\bibitem[Rothenh{\"a}usler et~al\mbox{.}(2018)]%
        {rothenhausler2018anchor}
\bibfield{author}{\bibinfo{person}{Dominik Rothenh{\"a}usler},
  \bibinfo{person}{Nicolai Meinshausen}, \bibinfo{person}{Peter B{\"u}hlmann},
  {and} \bibinfo{person}{Jonas Peters}.} \bibinfo{year}{2018}\natexlab{}.
\newblock \showarticletitle{Anchor regression: heterogeneous data meets
  causality}.
\newblock \bibinfo{journal}{\emph{arXiv preprint arXiv:1801.06229}}
  (\bibinfo{year}{2018}).
\newblock


\bibitem[Shapiro et~al\mbox{.}(2014)]%
        {shapiro2014lectures}
\bibfield{author}{\bibinfo{person}{Alexander Shapiro}, \bibinfo{person}{Darinka
  Dentcheva}, {and} \bibinfo{person}{Andrzej Ruszczy{\'n}ski}.}
  \bibinfo{year}{2014}\natexlab{}.
\newblock \bibinfo{booktitle}{\emph{Lectures on stochastic programming:
  modeling and theory}}.
\newblock \bibinfo{publisher}{SIAM}.
\newblock


\bibitem[Si et~al\mbox{.}(2020)]%
        {si2020distributional}
\bibfield{author}{\bibinfo{person}{Nian Si}, \bibinfo{person}{Fan Zhang},
  \bibinfo{person}{Zhengyuan Zhou}, {and} \bibinfo{person}{Jose Blanchet}.}
  \bibinfo{year}{2020}\natexlab{}.
\newblock \showarticletitle{Distributional Robust Batch Contextual Bandits}.
\newblock \bibinfo{journal}{\emph{arXiv preprint arXiv:2006.05630}}
  (\bibinfo{year}{2020}).
\newblock


\bibitem[Sinha et~al\mbox{.}(2018)]%
        {sinha2018certifying}
\bibfield{author}{\bibinfo{person}{Aman Sinha}, \bibinfo{person}{Hongseok
  Namkoong}, {and} \bibinfo{person}{John Duchi}.}
  \bibinfo{year}{2018}\natexlab{}.
\newblock \showarticletitle{Certifying Some Distributional Robustness with
  Principled Adversarial Training}. In \bibinfo{booktitle}{\emph{International
  Conference on Learning Representations}}.
\newblock


\bibitem[Staib and Jegelka(2017)]%
        {staib2017distributionally}
\bibfield{author}{\bibinfo{person}{Matthew Staib} {and}
  \bibinfo{person}{Stefanie Jegelka}.} \bibinfo{year}{2017}\natexlab{}.
\newblock \showarticletitle{Distributionally robust deep learning as a
  generalization of adversarial training}. In \bibinfo{booktitle}{\emph{NIPS
  workshop on Machine Learning and Computer Security}}.
\newblock


\bibitem[Subbaswamy et~al\mbox{.}(2021)]%
        {subbaswamy2021evaluating}
\bibfield{author}{\bibinfo{person}{Adarsh Subbaswamy}, \bibinfo{person}{Roy
  Adams}, {and} \bibinfo{person}{Suchi Saria}.}
  \bibinfo{year}{2021}\natexlab{}.
\newblock \showarticletitle{Evaluating Model Robustness and Stability to
  Dataset Shift}. In \bibinfo{booktitle}{\emph{Proceedings of The 24th
  International Conference on Artificial Intelligence and Statistics}}
  \emph{(\bibinfo{series}{Proceedings of Machine Learning Research},
  Vol.~\bibinfo{volume}{130})}, \bibfield{editor}{\bibinfo{person}{Arindam
  Banerjee} {and} \bibinfo{person}{Kenji Fukumizu}} (Eds.).
  \bibinfo{publisher}{PMLR}, \bibinfo{pages}{2611--2619}.
\newblock
\urldef\tempurl%
\url{http://proceedings.mlr.press/v130/subbaswamy21a.html}
\showURL{%
\tempurl}


\bibitem[Subbaswamy et~al\mbox{.}(2019)]%
        {subbaswamy2019preventing}
\bibfield{author}{\bibinfo{person}{Adarsh Subbaswamy}, \bibinfo{person}{Peter
  Schulam}, {and} \bibinfo{person}{Suchi Saria}.}
  \bibinfo{year}{2019}\natexlab{}.
\newblock \showarticletitle{Preventing failures due to dataset shift: Learning
  predictive models that transport}. In \bibinfo{booktitle}{\emph{The 22nd
  International Conference on Artificial Intelligence and Statistics}}.
  \bibinfo{pages}{3118--3127}.
\newblock


\bibitem[Sutton and Barto(2018)]%
        {sutton2018reinforcement}
\bibfield{author}{\bibinfo{person}{Richard~S Sutton} {and}
  \bibinfo{person}{Andrew~G Barto}.} \bibinfo{year}{2018}\natexlab{}.
\newblock \bibinfo{booktitle}{\emph{Reinforcement learning: An introduction}}.
\newblock \bibinfo{publisher}{MIT press}.
\newblock


\bibitem[Tamar et~al\mbox{.}(2014)]%
        {tamar2014scaling}
\bibfield{author}{\bibinfo{person}{Aviv Tamar}, \bibinfo{person}{Shie Mannor},
  {and} \bibinfo{person}{Huan Xu}.} \bibinfo{year}{2014}\natexlab{}.
\newblock \showarticletitle{Scaling up robust MDPs using function
  approximation}. In \bibinfo{booktitle}{\emph{International Conference on
  Machine Learning}}. \bibinfo{pages}{181--189}.
\newblock


\bibitem[Taori et~al\mbox{.}(2020)]%
        {taori2020measuring}
\bibfield{author}{\bibinfo{person}{Rohan Taori}, \bibinfo{person}{Achal Dave},
  \bibinfo{person}{Vaishaal Shankar}, \bibinfo{person}{Nicholas Carlini},
  \bibinfo{person}{Benjamin Recht}, {and} \bibinfo{person}{Ludwig Schmidt}.}
  \bibinfo{year}{2020}\natexlab{}.
\newblock \showarticletitle{Measuring Robustness to Natural Distribution Shifts
  in Image Classification}. In \bibinfo{booktitle}{\emph{Advances in Neural
  Information Processing Systems}},
  \bibfield{editor}{\bibinfo{person}{H.~Larochelle},
  \bibinfo{person}{M.~Ranzato}, \bibinfo{person}{R.~Hadsell},
  \bibinfo{person}{M.~F. Balcan}, {and} \bibinfo{person}{H.~Lin}} (Eds.),
  Vol.~\bibinfo{volume}{33}. \bibinfo{publisher}{Curran Associates, Inc.},
  \bibinfo{pages}{18583--18599}.
\newblock
\urldef\tempurl%
\url{https://proceedings.neurips.cc/paper/2020/file/d8330f857a17c53d217014ee776bfd50-Paper.pdf}
\showURL{%
\tempurl}


\bibitem[Thomas and Brunskill(2016)]%
        {thomas2016data}
\bibfield{author}{\bibinfo{person}{Philip Thomas} {and} \bibinfo{person}{Emma
  Brunskill}.} \bibinfo{year}{2016}\natexlab{}.
\newblock \showarticletitle{Data-efficient off-policy policy evaluation for
  reinforcement learning}. In \bibinfo{booktitle}{\emph{International
  Conference on Machine Learning}}. \bibinfo{pages}{2139--2148}.
\newblock


\bibitem[Uehara et~al\mbox{.}(2020)]%
        {kato2020off}
\bibfield{author}{\bibinfo{person}{Masatoshi Uehara}, \bibinfo{person}{Masahiro
  Kato}, {and} \bibinfo{person}{Shota Yasui}.} \bibinfo{year}{2020}\natexlab{}.
\newblock \showarticletitle{Off-Policy Evaluation and Learning for External
  Validity under a Covariate Shift}. In \bibinfo{booktitle}{\emph{Advances in
  Neural Information Processing Systems}},
  \bibfield{editor}{\bibinfo{person}{H.~Larochelle},
  \bibinfo{person}{M.~Ranzato}, \bibinfo{person}{R.~Hadsell},
  \bibinfo{person}{M.~F. Balcan}, {and} \bibinfo{person}{H.~Lin}} (Eds.),
  Vol.~\bibinfo{volume}{33}. \bibinfo{publisher}{Curran Associates, Inc.},
  \bibinfo{pages}{49--61}.
\newblock
\urldef\tempurl%
\url{https://proceedings.neurips.cc/paper/2020/file/0084ae4bc24c0795d1e6a4f58444d39b-Paper.pdf}
\showURL{%
\tempurl}


\bibitem[van Erven and Harremos(2014)]%
        {vanerven2014renyi}
\bibfield{author}{\bibinfo{person}{Tim van Erven} {and} \bibinfo{person}{Peter
  Harremos}.} \bibinfo{year}{2014}\natexlab{}.
\newblock \showarticletitle{Rényi Divergence and Kullback-Leibler Divergence}.
\newblock \bibinfo{journal}{\emph{IEEE Transactions on Information Theory}}
  \bibinfo{volume}{60}, \bibinfo{number}{7} (\bibinfo{year}{2014}),
  \bibinfo{pages}{3797--3820}.
\newblock
\urldef\tempurl%
\url{https://doi.org/10.1109/TIT.2014.2320500}
\showDOI{\tempurl}


\bibitem[Virtanen et~al\mbox{.}(2020)]%
        {scipy}
\bibfield{author}{\bibinfo{person}{Pauli Virtanen}, \bibinfo{person}{Ralf
  Gommers}, \bibinfo{person}{Travis~E. Oliphant}, \bibinfo{person}{Matt
  Haberland}, \bibinfo{person}{Tyler Reddy}, \bibinfo{person}{David
  Cournapeau}, \bibinfo{person}{Evgeni Burovski}, \bibinfo{person}{Pearu
  Peterson}, \bibinfo{person}{Warren Weckesser}, \bibinfo{person}{Jonathan
  Bright}, \bibinfo{person}{St{\'e}fan~J. {van der Walt}},
  \bibinfo{person}{Matthew Brett}, \bibinfo{person}{Joshua Wilson},
  \bibinfo{person}{K.~Jarrod Millman}, \bibinfo{person}{Nikolay Mayorov},
  \bibinfo{person}{Andrew R.~J. Nelson}, \bibinfo{person}{Eric Jones},
  \bibinfo{person}{Robert Kern}, \bibinfo{person}{Eric Larson},
  \bibinfo{person}{C~J Carey}, \bibinfo{person}{{\.I}lhan Polat},
  \bibinfo{person}{Yu Feng}, \bibinfo{person}{Eric~W. Moore},
  \bibinfo{person}{Jake {VanderPlas}}, \bibinfo{person}{Denis Laxalde},
  \bibinfo{person}{Josef Perktold}, \bibinfo{person}{Robert Cimrman},
  \bibinfo{person}{Ian Henriksen}, \bibinfo{person}{E.~A. Quintero},
  \bibinfo{person}{Charles~R. Harris}, \bibinfo{person}{Anne~M. Archibald},
  \bibinfo{person}{Ant{\^o}nio~H. Ribeiro}, \bibinfo{person}{Fabian Pedregosa},
  \bibinfo{person}{Paul {van Mulbregt}}, {and} \bibinfo{person}{{SciPy 1.0
  Contributors}}.} \bibinfo{year}{2020}\natexlab{}.
\newblock \showarticletitle{{{SciPy} 1.0: Fundamental Algorithms for Scientific
  Computing in Python}}.
\newblock \bibinfo{journal}{\emph{Nature Methods}}  \bibinfo{volume}{17}
  (\bibinfo{year}{2020}), \bibinfo{pages}{261--272}.
\newblock
\urldef\tempurl%
\url{https://doi.org/10.1038/s41592-019-0686-2}
\showDOI{\tempurl}


\bibitem[Wiesemann et~al\mbox{.}(2013)]%
        {wiesemann2013robust}
\bibfield{author}{\bibinfo{person}{Wolfram Wiesemann}, \bibinfo{person}{Daniel
  Kuhn}, {and} \bibinfo{person}{Berç Rustem}.}
  \bibinfo{year}{2013}\natexlab{}.
\newblock \showarticletitle{Robust Markov Decision Processes}.
\newblock \bibinfo{journal}{\emph{Mathematics of Operations Research}}
  \bibinfo{volume}{38}, \bibinfo{number}{1} (\bibinfo{year}{2013}),
  \bibinfo{pages}{153--183}.
\newblock
\urldef\tempurl%
\url{https://doi.org/10.1287/moor.1120.0566}
\showDOI{\tempurl}
\showeprint{https://doi.org/10.1287/moor.1120.0566}


\bibitem[Zhang et~al\mbox{.}(2020)]%
        {zhang2020invariant}
\bibfield{author}{\bibinfo{person}{Amy Zhang}, \bibinfo{person}{Clare Lyle},
  \bibinfo{person}{Shagun Sodhani}, \bibinfo{person}{Angelos Filos},
  \bibinfo{person}{Marta Kwiatkowska}, \bibinfo{person}{Joelle Pineau},
  \bibinfo{person}{Yarin Gal}, {and} \bibinfo{person}{Doina Precup}.}
  \bibinfo{year}{2020}\natexlab{}.
\newblock \showarticletitle{Invariant Causal Prediction for Block {MDP}s}. In
  \bibinfo{booktitle}{\emph{Proceedings of the 37th International Conference on
  Machine Learning}} \emph{(\bibinfo{series}{Proceedings of Machine Learning
  Research}, Vol.~\bibinfo{volume}{119})},
  \bibfield{editor}{\bibinfo{person}{Hal~Daumé III} {and}
  \bibinfo{person}{Aarti Singh}} (Eds.). \bibinfo{publisher}{PMLR},
  \bibinfo{pages}{11214--11224}.
\newblock
\urldef\tempurl%
\url{http://proceedings.mlr.press/v119/zhang20t.html}
\showURL{%
\tempurl}


\bibitem[Zhou et~al\mbox{.}(2021)]%
        {zhou2021tabular}
\bibfield{author}{\bibinfo{person}{Zhengqing Zhou}, \bibinfo{person}{Zhengyuan
  Zhou}, \bibinfo{person}{Qinxun Bai}, \bibinfo{person}{Linhai Qiu},
  \bibinfo{person}{Jose Blanchet}, {and} \bibinfo{person}{Peter Glynn}.}
  \bibinfo{year}{2021}\natexlab{}.
\newblock \showarticletitle{Finite-Sample Regret Bound for Distributionally
  Robust Offline Tabular Reinforcement Learning}. In
  \bibinfo{booktitle}{\emph{Proceedings of The 24th International Conference on
  Artificial Intelligence and Statistics}} \emph{(\bibinfo{series}{Proceedings
  of Machine Learning Research}, Vol.~\bibinfo{volume}{130})},
  \bibfield{editor}{\bibinfo{person}{Arindam Banerjee} {and}
  \bibinfo{person}{Kenji Fukumizu}} (Eds.). \bibinfo{publisher}{PMLR},
  \bibinfo{pages}{3331--3339}.
\newblock
\urldef\tempurl%
\url{https://proceedings.mlr.press/v130/zhou21d.html}
\showURL{%
\tempurl}


\end{thebibliography}

\appendix

\onecolumn
\appendix

\section{Implementation details of solving DRO given a finite sample}\label{app:dro_bound}
The DRO problem for $\cU_P^{\text{div}}$ solves,
\begin{equation}
\label{eq:robustint_app}
\argmin_{\theta\in \Theta} \left\{\R(\theta, \cU_P^{\text{div}}):= \sup_{Q\in\ \cU_P^{\text{div}}}\ \E_{\V\sim Q}[\ell(\theta, \V)]\right\}
\end{equation}
Using convex duality arguments \citep{shapiro2014lectures}, one can write the worst-case risk alternatively as,
\begin{equation}
\label{eq:robust_marginal}
\begin{aligned}
    \R(\theta, \cU_P^{\text{div}}) &= \sup_{Q\in\ \cU_P^{\text{div}}}\ \E_{Z\sim Q(Z)}\E_{P(V\vert Z)}[\ell(\theta,V)\vert Z] = \inf_{\eta\in \reals} \frac{1}{\delta}\E_P\left[(\E_P\left[\ell(\theta,V)\vert Z\right]-\eta)_+\right] + \eta
\end{aligned}
\end{equation}
where $(\cdot)_+ = \text{max}(\cdot, 0)$. 
Note that this requires estimating $\E_{P(V\vert Z)}[\cdot]$ which may be hard if $Z$ is continuous-valued or we do not have enough data for all possible values of $Z$. 
Assuming smoothness of this conditional loss, \citet[][Lemma 4.2]{duchi2019distributionally} gives an upper bound for the worst-case risk with the empirical version as, 

\begin{align}
\begin{split}
    \label{eq:marginal_smooth_app}
    \widehat{\R}(\theta, \cU_P^{\text{div}})
    = \inf_{\substack{\eta\geq 0,\\ B\in \reals^{n\times n}_+}}  %
    \Bigg\{& \frac{1}{\delta} \Big(\frac{1}{n}\sum_{i=1}^n\Big(\ell(\theta, V_i) {-} \frac{1}{n}\sum_{j=1}^n(B_{ij}{-}B_{ji}) {-} \eta\Big)^2_+\Big)^{\frac{1}{2}} %
    + \frac{L}{\epsilon n^2}\sum_{i,j=1}^n \sum_{O\in Z}\Vert O_i-O_j\Vert B_{ij} + \eta \Bigg\} %
    \end{split}
\end{align}

for any $\epsilon>0$, where $\eta, B$ are dual variables. We use this estimator while solving the DRO problems in the contextual bandit experiments. The minimization is performed using gradient descent as the objective is convex in both $\eta$ and $B$.

\section{Assumptions for MDPs -- SA-Rectangularity}
\label{app:assum_mdp}

We assume that the uncertainty sets are SA-rectangular \citep{iyengar2005robust}, which intuitively implies that the uncertainty sets are constructed across time steps independently. This property yields a tractable method to compute the value function estimates using dynamic programming \citep{iyengar2005robust}.

\begin{definition}[SA-rectangularity]
For any state-action pair (s,a), consider the uncertainty set $\U^{\text{MDP}}(s,a)$ of plausible test-time transition models. Denote the collection of all such uncertainty sets by  $\U^{\text{MDP}} \ni \U^{\text{MDP}}(s,a)$. Then, SA-rectangularity means that the collection $\U^{\text{MDP}}$ is constructed by taking the Cartesian product of the individual sets $\U^{\text{MDP}}(s,a)$. 
$$\U^{\text{MDP}} = \bigtimes_{(s,a)\in\mathcal{S}\times\mathcal{A}} \U^{\text{MDP}}(s,a)$$
\end{definition}
For example, take an MDP with one state and two actions, $\mathcal{S}=\{0\}$, $\mathcal{A}=\{0,1\}$. If $\{\{P_{00}^1,P_{00}^2\},\{P_{01}^1,P_{01}^2\}\}$ are the two uncertainty sets for the two state-action pairs $\{00,01\}$ respectively, then the four possible uncertainty sets for the robust MDP are $\{\{P_{00}^1,P_{01}^1\},\{P_{00}^1,P_{01}^2\},\{P_{00}^2,P_{01}^1\},\{P_{00}^2,P_{01}^2\}\}$.

We note that some policy evaluation methods for robust MDPs require weaker assumptions than SA-rectangularity \citep{wiesemann2013robust}. %

\section{Proof of Theorem \ref{thm:error_ope} -- error bound for robust OPE with estimated model}
\label{app:error_ope}

We solve the robust OPE problem using dynamic programming with the \textit{estimated} transition and reward models instead of assuming access to the true models. Hence, we will incur an error due to using finite samples to estimate the models and the value. Here, we will show that this error is of the order of, as in the non-robust OPE case, $\tilde{O}(\frac{\gamma |\mathcal{S}|}{\sqrt{n}(1-\gamma)^2})$ (ignoring logarithmic factors).

\textbf{Notation}.
The transition model for the training environment is written as $P_0$. We will abbreviate the uncertainty set defined with respect to $P_0$ for each state-action pair, that is $\mathcal{U}_{P_0}(s,a)$, as $\mathcal{U}_{P_0}$. For finite state-action space, a policy's value $V^\pi$ is a $|\mathcal{S}|$ dimensional vector. Thus, the $L_\infty$ norm for this Banach space is $\|V^\pi\|_\infty = \max_s |V^\pi(s)|$.
The robust Bellman operator is defined as 
$\mathcal{T}_{\mathcal{U}_{P_0}} V^\pi = r + \inf_{P\in\mathcal{U}_{P_0}} \gamma \langle P, V^\pi\rangle$.
The reward and transition models are denoted by $M:=(r_0, P_0)$, which are $|\mathcal{S}|\times|\mathcal{A}|$ and $|\mathcal{S}|\times|\mathcal{A}\times|\mathcal{S}|$ tensors respectively. Denote the estimated models by $\widehat{M}_0:=(\widehat{r}_0, \widehat{P}_0)$. The robust value for policy $\pi$ and state $s$ is denoted by $V^\pi_{\mathcal{U}_{P_0}}(s)$. We exclude $\widehat{r}_0$ to avoid excessive notation. When this value is computed instead with the estimated models $\widehat{M}_0$, as done in our experiments, we will denote the value by $V^\pi_{\mathcal{U}_{\widehat{P}_0}}(s)$. 

Thus, our goal is to bound the error between $V^\pi_{\mathcal{U}_{P_0}}(s)$ and $V^\pi_{\mathcal{U}_{\widehat{P}_0}}(s)$. For the non-robust MDP case, this result is referred to as \textit{simulation lemma} \citep{kearns2002near}. It bounds the error in value estimates incurred from using the estimated models instead of the true ones. An alternative statement of the lemma and an accessible proof is provided in \citep[][Lemma 1]{jiang2020notes}. In Lemma \ref{lem:simulation}, we will first bound the difference between $V^\pi_{\mathcal{U}_{P_0}}(s)$ and $V^\pi_{\mathcal{U}_{\widehat{P}_0}}(s)$ assuming that the estimation error for the reward and transition models is bounded. Then, we will use standard concentration inequalities to prove the error bounds. Before proving the result, we make some observations required in the proof.

\textbf{Remark 1 (Upper bound of value)}.
Assume that the rewards in the MDP are bounded, $r\in[0,r_\text{max}]$. Then, the robust value is upper bounded by $r_\text{max}/(1-\gamma)$. This follows from the definition of the robust value as the infimum,
\begin{align*}
    V^\pi_{\mathcal{U}_{P_0}}(s) &= \inf_{P\in\mathcal{U}_{P_0}} \E_{\pi,P}\left[\sum_{t=0}^\infty \gamma^t r(s_t,a_t)\vert s_0\right]
    \leq \E_{\pi,P_0}\left[\sum_{t=0}^\infty \gamma^t r(s_t,a_t)\vert s_0\right] \leq \sum_{t=0}^\infty \gamma^t r_\text{max} = \frac{r_\text{max}}{1-\gamma}
\end{align*}

\textbf{Remark 2 (Contraction property)}. From \citet[][Theorem 5]{iyengar2005robust}, we know that the robust Bellman operator $\mathcal{T}_{\mathcal{U}_{P_0}}$ is a \textit{contraction mapping} in $L_\infty$ norm. That is, for any two value functions $U$ and $V$, 
\begin{align}
\label{eq:contraction}
    \|\mathcal{T}_{\mathcal{U}_{P_0}} V - \mathcal{T}_{\mathcal{U}_{P_0}} U\|_\infty \leq \gamma \|V-U\|_\infty
\end{align}

Now, we prove a lemma from which the error bound follows directly.
\begin{lemma}[Simulation lemma for robust MDPs]
\label{lem:simulation}
Given $\max_{s,a} |r(s,a)-\widehat{r}(s,a)| \leq \epsilon_r$ and $\max_{s,a} \|P_0(\cdot|s,a)-\widehat{P}_0(\cdot|s,a)\|_1 \leq \epsilon_P$, then for any policy $\pi$, 
\begin{align*}
    \|V^\pi_{\mathcal{U}_{P_0}} - V^\pi_{\mathcal{U}_{\widehat{P}_0}}\|_\infty \leq \frac{\epsilon_r}{1-\gamma} + \frac{\gamma\epsilon_P r_\text{max}}{(1-\gamma)^2}
\end{align*}
\end{lemma}
\begin{proof}
The proof follows the exposition of the simulation lemma by \citet{jiang2020notes} for non-robust MDPs.
The steps involve using the robust Bellman operator, then noting that it is a contraction, and finally, using the fact that robust value which is defined as infimum over the uncertainty set is lower than the value under the training transition model.

Recall that the robust value with respect to an uncertainty set defined using $P_0$ satisfies the following robust Bellman equation,
$$V^\pi_{\mathcal{U}_{P_0}}(s) = \sum_a \pi(a|s)\left(r(s,a) + \inf_{P\in\mathcal{U}_{P_0}} \gamma \langle P, V^\pi_{\mathcal{U}_{P_0}}\rangle\right)$$
Here, the policy $\pi$ can be either probabilistic or deterministic, in which case $\pi(a|s)=1$ for one of the actions $a$.

For all $s\in \mathcal{S}$, consider the difference between robust values defined with respect to $P_0$ and $\widehat{P}_0$,
\begin{align}
    \Big{|}V^\pi_{\mathcal{U}_{P_0}}(s) - V^\pi_{\mathcal{U}_{\widehat{P}_0}}(s)\Big{|} &= \Big{|}\sum_a \pi(a|s)(r(s,a) + \inf_{P\in\mathcal{U}_{P_0}} \gamma \langle P, V^\pi_{\mathcal{U}_{P_0}}\rangle) - \sum_a \pi(a|s)(\widehat{r}(s,a) + \inf_{P\in\mathcal{U}_{\widehat{P}_0}} \gamma \langle P, V^\pi_{\mathcal{U}_{\widehat{P}_0}}\rangle)\Big{|}\nonumber\\
    &\leq \Big{|}\sum_a \pi(a|s)(r(s,a) - \widehat{r}(s,a))\Big{|} + \Big{|}\sum_a \pi(a|s) (\inf_{P\in\mathcal{U}_{P_0}} \gamma \langle P, V^\pi_{\mathcal{U}_{P_0}}\rangle - \inf_{P\in\mathcal{U}_{\widehat{P}_0}} \gamma \langle P, V^\pi_{\mathcal{U}_{\widehat{P}_0}}\rangle)\Big{|}\nonumber\\
    &\leq \epsilon_r + \sum_a \pi(a|s) \underbrace{\Big{|}\inf_{P\in\mathcal{U}_{P_0}} \gamma \langle P, V^\pi_{\mathcal{U}_{P_0}}\rangle - \inf_{P\in\mathcal{U}_{\widehat{P}_0}} \gamma \langle P, V^\pi_{\mathcal{U}_{\widehat{P}_0}}\rangle\Big{|}}_{\text{Part (I)}}\label{eq:simulation_infval}
\end{align}
where we used the bound on estimated rewards from the assumption. We now bound Part (I) for each pair $(s,a)$.
\begin{align}
    \text{Part (I)} &\leq |(\inf_{P\in\mathcal{U}_{P_0}} \gamma \langle P, V^\pi_{\mathcal{U}_{P_0}}\rangle - \inf_{P\in\mathcal{U}_{P_0}} \gamma \langle P, V^\pi_{\mathcal{U}_{\widehat{P}_0}}\rangle) + (\inf_{P\in\mathcal{U}_{P_0}} \gamma \langle P, V^\pi_{\mathcal{U}_{\widehat{P}_0}}\rangle - \inf_{P\in\mathcal{U}_{\widehat{P}_0}} \gamma \langle P, V^\pi_{\mathcal{U}_{\widehat{P}_0}}\rangle)|\nonumber\\
    &= |(\mathcal{T}_{\mathcal{U}_{P_0}} V^\pi_{\mathcal{U}_{P_0}} - \mathcal{T}_{\mathcal{U}_{P_0}} V^\pi_{\mathcal{U}_{\widehat{P}_0}}) + (\inf_{P\in\mathcal{U}_{P_0}} \gamma \langle P, V^\pi_{\mathcal{U}_{\widehat{P}_0}}\rangle - \inf_{P\in\mathcal{U}_{\widehat{P}_0}} \gamma \langle P, V^\pi_{\mathcal{U}_{\widehat{P}_0}}\rangle)|\nonumber\\
    &\leq \gamma\|V^\pi_{\mathcal{U}_{P_0}} - V^\pi_{\mathcal{U}_{\widehat{P}_0}}\|_\infty + |(\inf_{P\in\mathcal{U}_{P_0}} \gamma \langle P, V^\pi_{\mathcal{U}_{\widehat{P}_0}}\rangle - \inf_{P\in\mathcal{U}_{\widehat{P}_0}} \gamma \langle P, V^\pi_{\mathcal{U}_{\widehat{P}_0}}\rangle)|\label{eq:simulation_contraction}\\
    &\leq \gamma\|V^\pi_{\mathcal{U}_{P_0}} - V^\pi_{\mathcal{U}_{\widehat{P}_0}}\|_\infty + \underbrace{|(\gamma \langle P_0, V^\pi_{\mathcal{U}_{\widehat{P}_0}}\rangle - \inf_{P\in\mathcal{U}_{\widehat{P}_0}} \gamma \langle P, V^\pi_{\mathcal{U}_{\widehat{P}_0}}\rangle)|}_\text{Part (II)}\label{eq:simulation_infstep}\\
\end{align}
where we used the contraction property of $\mathcal{T}_{\mathcal{U}_{P_0}}$ from Remark 2 in (\ref{eq:simulation_contraction}) and replaced $\inf$ with a member from the set in (\ref{eq:simulation_infstep}).

Using the H\"{o}lder's inequality and the upper bound from Remark 1, we can write Part (II) as,
\begin{align*}
    \text{Part (II)} = |\sup_{P\in\mathcal{U}_{\widehat{P}_0}} \gamma \langle P_0 - P, V^\pi_{\mathcal{U}_{\widehat{P}_0}}\rangle| &\leq \gamma \sup_{P\in\mathcal{U}_{\widehat{P}_0}} |\langle P_0 - P, V^\pi_{\mathcal{U}_{\widehat{P}_0}}\rangle|\\
    &\leq \gamma \sup_{P\in\mathcal{U}_{\widehat{P}_0}} \|P_0 - P\|_1 \|V^\pi_{\mathcal{U}_{\widehat{P}_0}}\|_\infty\\
    &= \gamma \sup_{P\in\mathcal{U}_{\widehat{P}_0}} \|P_0 - P\|_1 \frac{r_\text{max}}{1-\gamma} \leq \gamma \sup_{P\in\mathcal{U}_{\widehat{P}_0}} (\|P_0 - \widehat{P}_0 \|_1 + \|\widehat{P}_0 - P\|_1) \frac{r_\text{max}}{1-\gamma}
\end{align*}
We re-write the $\ell_1$ norm in terms of total variation distance (TV) \citep[][Prop 4.2]{levin2017markov} and use Pinsker's inequality \citep[][Lemma 2]{canonne2022note},
\begin{align*}
    \text{Part (II)} &\leq \gamma \epsilon_P \frac{r_\text{max}}{1-\gamma} + \gamma \sup_{P\in\mathcal{U}_{\widehat{P}_0}} 2\ \text{TV}(\widehat{P}_0, P) \frac{r_\text{max}}{1-\gamma}
    \leq \gamma \epsilon_P \frac{r_\text{max}}{1-\gamma} + \gamma \sup_{P\in\mathcal{U}_{\widehat{P}_0}} 2\ \sqrt{\frac{1}{2}\text{KL}(\widehat{P}_0, P)} \frac{r_\text{max}}{1-\gamma}
\end{align*}
By the definition of the uncertainty set $\mathcal{U}_{\widehat{P}_0}$, all elements $P$ satisfy $\text{KL}(\widehat{P}_0, P) \leq \delta$.
\begin{align*}
    \text{Part (II)} \leq \gamma \epsilon_P \frac{r_\text{max}}{1-\gamma} + \gamma \sqrt{2\delta} \frac{r_\text{max}}{1-\gamma} = \gamma (\epsilon_P + \sqrt{2\delta}) \frac{r_\text{max}}{1-\gamma} =: \gamma (\epsilon'_P) \frac{r_\text{max}}{1-\gamma}
\end{align*}
where we define $\epsilon'_P := \epsilon_P + \sqrt{2\delta}$. Now, we plug this back in (\ref{eq:simulation_infstep}) and use $\sum_a \pi(a|s)=1$ in (\ref{eq:simulation_infval}) to get,
$$\Big{|}V^\pi_{\mathcal{U}_{P_0}}(s) - V^\pi_{\mathcal{U}_{\widehat{P}_0}}(s)\Big{|} \leq \epsilon_r + \gamma\|V^\pi_{\mathcal{U}_{P_0}} - V^\pi_{\mathcal{U}_{\widehat{P}_0}}\|_\infty + \gamma \epsilon'_P \frac{r_\text{max}}{1-\gamma}$$

Since the above holds for any $s\in\mathcal{S}$, taking maximum over $\mathcal{S}$, we have that
\begin{align*}
    \|V^\pi_{\mathcal{U}_{P_0}} - V^\pi_{\mathcal{U}_{\widehat{P}_0}}\|_\infty &\leq \epsilon_r + \gamma\|V^\pi_{\mathcal{U}_{P_0}} - V^\pi_{\mathcal{U}_{\widehat{P}_0}}\|_\infty + \gamma \epsilon'_P \frac{r_\text{max}}{1-\gamma}\\
    \|V^\pi_{\mathcal{U}_{P_0}} - V^\pi_{\mathcal{U}_{\widehat{P}_0}}\|_\infty &\leq \frac{\epsilon_r}{1-\gamma} + \frac{\gamma\epsilon'_P r_\text{max}}{(1-\gamma)^2}
\end{align*}
\end{proof}

Restating the result that we want to prove,
\begin{theorem*}[Estimation error for robust OPE]
Given at least $n$ samples from $P_0(\cdot\vert s,a)$ for all $s,a$, assuming that the rewards are bounded $r\in[0,r_\text{max}]$, and that $\mathcal{U}^{\text{MDP}}$ are defined by KL-divergence, then with probability at least $1-\alpha$,
\begin{align*}
    \|V^\pi_{\mathcal{U}_{P_0}} - V^\pi_{\mathcal{U}_{\widehat{P}_0}}\|_\infty \leq O\left(\frac{\gamma r_\text{max}|\mathcal{S}|}{(1-\gamma)^2} \sqrt{\frac{1}{n}\log\left(\frac{4|\mathcal{S}\times\mathcal{A}\times\mathcal{S}|}{\alpha}\right)}\right)
\end{align*}
\end{theorem*}
\begin{proof}
We use Lemma \ref{lem:simulation} along with Hoeffding's inequality \citep{hoeffding1994probability} to bound the error in terms of number of samples and state-action pairs.

Say, we have $n$ samples to estimate $\widehat{r}_0(s,a),\widehat{P}_0(s,a)$. Then, after using Hoeffding's inequality for each state-action pair and then with the union bound, we have that with probability at least $1-\alpha$,
\begin{align*}
    \max_{s,a} |r(s,a)-\widehat{r}(s,a)| \leq r_\text{max}\sqrt{\frac{1}{2n}\log\left(\frac{4|\mathcal{S}\times\mathcal{A}|}{\alpha}\right)},\qquad \quad
    &\max_{s,a,s'} |P_0(s'|s,a)-\widehat{P}_0(s'|s,a)| \leq \sqrt{\frac{1}{2n}\log\left(\frac{4|\mathcal{S}\times\mathcal{A}\times\mathcal{S}|}{\alpha}\right)},\\
    \implies &\max_{s,a} \|P_0(\cdot|s,a)-\widehat{P}_0(\cdot|s,a)\|_1 \leq |\mathcal{S}| \sqrt{\frac{1}{2n}\log\left(\frac{4|\mathcal{S}\times\mathcal{A}\times\mathcal{S}|}{\alpha}\right)}.
\end{align*}
Thus, $\|V^\pi_{\mathcal{U}_{P_0}} - V^\pi_{\mathcal{U}_{\widehat{P}_0}}\|_\infty = \tilde{O}(\frac{\gamma |\mathcal{S}|}{\sqrt{n}(1-\gamma)^2})$.
\end{proof}
The dependence on $1/\sqrt{n}$ matches that in the non-robust case.
The dependence on $|\mathcal{S}|$ can be improved to $\sqrt{|\mathcal{S}|}$ using a better bound for $\max_{s,a,s'} |P_0(s'|s,a)-\widehat{P}_0(s'|s,a)|$, as done in \citep[][Section 2.2]{jiang2020notes}. We hope that further work will tighten this bound using techniques from \citep{li2020breaking}.
For Lemma \ref{lem:simulation}, we restricted to KL-divergence in order to use the Pinsker's inequality to bound the $\ell_1$ distance $\|P_0 - P\|_1$. We are not aware of a Pinsker-type inequality for our preferred divergence measure $\divcvar$ (while one exists for  R\'enyi divergences of order $\in (0,1]$, but not for order $\infty$ which is the same as $\divcvar$ \citep{vanerven2014renyi}). While the proposed method for MDPs works for KL-divergence as well, we use $\divcvar$ in the experiments due to its intuitive interpretation in terms of worst-case subpopulations.

\section{Algorithm for robust OPE for MDPs}
\label{app:algmdp}
Robust OPE with dynamic programming amounts to solving the following fixed point equation iteratively,
\begin{equation}
    \begin{aligned}
        V^\pi(s) = r(s,\pi(s)) + \inf_{P\in\U^{\text{MDP}}(s,\pi(s))} \gamma \E_{s'\sim P(s'\vert s,\pi(s))}[V^\pi(s')]
    \end{aligned}
\end{equation}

At each iteration, we have to solve the DRO problem,
\begin{align*}
    \inf_{P\in\mathcal{U}^{\text{MDP}}(s,\pi(s))} \E_P[V^\pi(s')]
    &= \inf_{P\in\mathcal{U}^{\text{MDP}}(s,\pi(s))} \E_{P(s'^{1}\vert s, \pi(s))} \left[\E_{P_0(s'^{2}\vert s, \pi(s), s'^{1})}\left[V^\pi(s')\right]\right] =: \cR_{\mathcal{U}^{\text{MDP}}(s, \pi(s))} \left(V^\pi(s')\right)
\end{align*}
We estimate the inner expectation w.r.t. $P_0$ with Monte-Carlo averaging on the batch data available to us as this data contains samples from $P_0$. We then compute the value function update using 
\begin{align}
    V^\pi(s) &\leftarrow \sum_a \pi(a\vert s) \left(r(s,a) + \gamma \cR_{\mathcal{U}^{\text{MDP}}(s, a)} \left(V^\pi(s')\right)\right). \label{eq:robust_mdp_app}
\end{align}
As earlier, we choose the divergence metric needed to define $\mathcal{U}^{\text{MDP}}(s, a)$ to be $\divcvar$, and accordingly, solve the DRO problem using the form in Eq. (\ref{eq:robust_marginal}). This requires access to the true transition model, $P_0$, and the reward model, $r$. Since we only have samples from $P_0$, we use the maximum likelihood estimate $\widehat{P_0}$ from the batch data in place of $P_0$. For finite state-action space, this amounts to counting transitions for each tuple $(s,a,s')$ and averaging. If we do not have access to the true or estimated reward model $r(s,a)$, we can use importance weighting to estimate the first term of Eq. (\ref{eq:robust_mdp_app}) as $\frac{\pi(a|s)}{\mu(a|s)}r(s,a)$ for the observed $(s,a,r(s,a))$ tuple. Here, an important requirement for the batch data  collected using policy $\mu$ is that it has sufficient exploration to evaluate the given policy $\pi$. Formally, $\pi(a|s)>0 \implies \mu(a|s)>0$ for all $s,a$. In words, the support of the batch policy $\mu$ contains the support of evaluation policy $\pi$. Further, the transition models in $\mathcal{U}^{\text{MDP}}(s, a)$ that we will have to evaluate are also absolutely continuous with respect to the training transition model $P_0$, as assumed in the set definition (\ref{eq:int_set}). Thus, given enough samples in the batch data collected using $\mu$ and $P_0$, we can estimate the reward and transition models for all $s,a,s'$ needed in the value function updates for $\pi$ and $\mathcal{U}^{\text{MDP}}$ in (\ref{eq:robust_mdp_app}).
Algorithm \ref{alg:cb_mdp} gives the overall procedure in detail.

\textbf{Extension to continuous state space}. Although we specify the method for finite state MDPs, it can be extended to continuous or large state spaces by leveraging linear function approximations along with robust dynamic programming, as proposed in \citep{tamar2014scaling}. Validating such approximations for $\ouralg$ is a fruitful direction for more work.

\section{Experimental details}\label{app:expts_sl}

\subsection{More detail on feature shift detection}
\label{app:shift_detect_app}
Details of the three datasets used in Figure \ref{fig:count_shifts} and the process followed for feature extraction is given in Adult Income \citep{ding2021retiring}, SBA \citep{min2018sba}, and eICU \citep{johnson2018generalizability}. We use the score-based feature shift detection method (named MB-SM) by \citep{kulinski2020feature} available at \url{https://github.com/inouye-lab/feature-shift}.

\subsection{Choosing the uncertainty set size $\delta$}
As our problem setup demands robustness to \textit{unknown} shifts at the test time, the choice of uncertainty set size (i.e. robustness level $\delta$) and other hyperparameters necessarily involves making assumptions about future data. For the case of $\divcvar$ based sets, the set size corresponds to the minimum possible proportion of the worst-performing subpopulation. The modeler can exercise domain knowledge in choosing such a proportion, for instance, using the summary statistics of demographic data for the intended test populations.

For example, if the modeler worries about shifts in the age distribution. Say the train distribution has 50-50\% young-old population, and at the worst the test distribution has 80\% overlap, i.e. old population is at the least 40\% (=0.8*50\%) at test time, then $\delta=0.8$ is reasonable to guarantee robustness. Summary statistics of demographics for the test environments (age, gender, comorbidity distribution at test hospitals) may help judge the size of the population in attributes of concern.

\subsection{Data generating process for synthetic data in CBs}
\label{app:data_cb}
We generate data according to the causal graph in Figure \ref{fig:model_cb} with two features $Z:=(Z_1, Z_2)$ in the context, binary treatment $T$ and continuous outcome $Y$. Structural equations are:
{\small{
\begin{align*}
& E\sim {\small{\text{Bernoulli}}}(-1,1,\delta_0),%
 Z_1\sim E\cdot \cN(10,1), 
 Z_2\sim \cN(5,1) \qquad
    Y_{T=t}\sim \cN(Z^\top w_t,{0.1}^2), t\in\{0,1\}%
     \text{, where } w_0=[0.1,0.1], w_1=[0.1,0.5] \\
    &T\sim \text{Bernoulli}(\sigma(Z^\top\beta + \beta_0)), \beta=[0.1,0.1]
\end{align*}}}
where $\sigma(z)=1/(1+e^{-z})$. The bias term in the train policy is $\beta_0=-1$, whereas the policy to be evaluated in test environments has $\beta_0=-0.5$. In addition, the marginal distribution of $Z_1$ is changed, via change in $E$, by increasing $\delta_0$ in the test environment. We simulate $n=2000$ samples in the train environment and use it for all methods. The objective is to estimate the robust value of the new policy. 

\subsection{More details on Cliffwalking domain}
\label{app:data_cliff}
We consider a $6\times6$ gridworld (Figure \ref{fig:cliff}) with start, goal positions, and a cliff on one edge \citep[][Ex. 6.6]{sutton2018reinforcement}. The agent incurs a rewards $-100$ on falling off the cliff, $-1$ for each step, and $0$ on reaching the goal.
To add stochasticity in transitions, we make two changes to the domain. With a constant shift probability, the agent slips down one step towards the cliff instead of taking the prescribed action. This shift probability varies across environments, thus, changing the transition dynamics, necessitating robustness in policy evaluation.
Second, we consider two state features in the observation space of the agent. The first feature is the agent's position on the grid, and the second is discrete random noise sampled uniformly from the set $\{1,2,\dots,6\times6\}$. The second feature has an associated reward sampled from Uniform$(-1,0)$ for the $6\times 6$ values it takes. Total reward is the sum of rewards from the two features (one based on grid position and one sampled uniformly between -1,0). 

Since agent's actions affect only the first feature, $\ouralg$ correctly constructs uncertainty sets based on transition dynamics in the first feature alone, $P(s^1\vert s,a)$. In contrast, $\divalg$ ignores this structure and constructs uncertainty sets using both the features, $P((s^1,s^2)\vert s,a)$.
We evaluate the value function estimate for an agent following uniform random policy using dynamic programming with the standard Bellman equation ($\stdalg$) or the robust one in Eq. (\ref{eq:robust_bellman}) ($\divalg,\ouralg$).

\subsection{More details on Sepsis domain}
The RL policy to be evaluated is obtained as follows. The optimal policy for the Sepsis simulator, namely the physician policy, is found with the procedure used in \citep{oberst19counterfactual}. Physician policy is made stochastic by taking random action with probability 0.05. With this policy, a dataset with $1000$ trajectories each with maximum length of $20$ is sampled. Diabetic population is fixed to 20\% while sampling. The RL policy used in evaluation is obtained by running policy iteration on this dataset. Then, RL policy is used to sample another dataset with $10000$ trajectories each with maximum length of $20$, and 20\% diabetics. This data is used for the final evaluation. As the policy being evaluated is the same as the one used to sample the available data, importance weighting of the value function updates in Eq. (\ref{eq:robust_mdp_app}) by the factor $\frac{\pi(a|s)}{\mu(a|s)}(=1)$ is not needed.

\end{document}